\documentclass[11pt]{article}

\usepackage[margin=1in]{geometry}
\usepackage{amsmath,amssymb}
\usepackage{array}
\usepackage{booktabs,longtable,multirow}
\usepackage{caption,subcaption}
\usepackage{changepage}
\usepackage{float}
\usepackage{graphicx}
\usepackage{textcomp}
\usepackage[dvipsnames,table]{xcolor}
\usepackage{microtype}
\usepackage[separate-uncertainty=true,multi-part-units=single]{siunitx}
\usepackage[numbers,sort&compress]{natbib}
\usepackage[hidelinks]{hyperref}
\usepackage[nameinlink,noabbrev]{cleveref}

\graphicspath{{Figures/}}
\newcommand{\pymc}{PyMC}

\begin{document}

\begin{center}
{\LARGE\bfseries Simulation-based inference for rapid Bayesian parameter estimation in epidemiological models: a comparison with MCMC\par}
\vspace{1em}
Alina Bazarova$^{1,2,*,\dagger}$,
Johann Fredrik Jadebeck$^{3,6,\dagger}$,
Henrik Zunker$^{4}$,
Carolina J. Klett-Tammen$^{5}$,
Torben Heinsohn$^{5}$,
Wolfgang Wiechert$^{3,6}$,
Katharina Nöh$^{3}$,
and Stefan Kesselheim$^{1,2}$
\vspace{0.75em}

{\small
$^1$ Forschungszentrum Jülich, Jülich Supercomputing Centre, Jülich, Germany\\
$^2$ Helmholtz AI, Germany\\
$^3$ Forschungszentrum Jülich, Institute of Bio- and Geosciences, Jülich, Germany\\
$^4$ German Aerospace Center, Institute of Software Technology, Department High-Performance Computing, Cologne, Germany\\
$^5$ Helmholtz Centre for Infection Research, Braunschweig, Germany\\
$^6$ RWTH Aachen University, Computational Systems Biology, Aachen, Germany\\[0.5em]
$^*$ Corresponding author: \href{mailto:al.bazarova@fz-juelich.de}{al.bazarova@fz-juelich.de}\\
$^\dagger$ These authors contributed equally to this work.
}
\end{center}
\vspace{1em}

\begin{abstract}

Mechanistic epidemiological models are widely used to support infectious disease forecasting and public-health decision making. Bayesian calibration of such models is commonly performed using Markov chain Monte Carlo (MCMC), which can become computationally expensive for high-dimensional nonlinear systems and repeated near-real-time analyses. Here, we investigate simulation-based inference (SBI) using neural posterior estimation as a scalable alternative for Bayesian calibration of a mechanistic SECIR epidemiological model using COVID-19 intensive care unit (ICU) occupancy data from Germany during 2020. We compared SBI and MCMC across multiple epidemic phases using both 31-day inference windows and a substantially more challenging 201-day reconstruction problem involving multiple transmission change points. Posterior agreement was evaluated quantitatively using Wasserstein distances and Kullback–Leibler divergences together with posterior predictive checks. Across the 31-day windows, SBI recovered posterior distributions in strong agreement with MCMC while accurately reproducing observed ICU trajectories. In the 201-day setting, SBI preserved the dominant posterior structure despite increased uncertainty. SBI, by combining CPU and GPU resources, substantially reduced computational runtime compared with MCMC, which was restricted to running on CPUs. Whereas MCMC required approximately 1000 seconds for the 31-day inference problems, SBI achieved comparable posterior and predictive performance in approximately 60–70 seconds on a single NVIDIA A100 GPU.
For the 201-day inference problem, SBI required an average of 157 seconds, while the MCMC
runs took over 19,000 seconds.
Our results demonstrate that SBI provides a rapid and computationally efficient framework for Bayesian calibration of mechanistic epidemiological models, supporting repeated near-real-time inference and rapid outbreak analysis.

\end{abstract}

\section*{Author summary}

Mechanistic epidemiological models played an important role during the COVID-19 pandemic by helping estimate disease transmission, forecast healthcare demand, and evaluate intervention strategies. These models require repeated calibration to surveillance data as epidemic conditions change. Bayesian inference provides a principled framework for such calibration while quantifying uncertainty, but standard approaches based on Markov chain Monte Carlo (MCMC) are often computationally expensive and difficult to apply in near-real-time settings.

In this study, we investigated simulation-based inference (SBI), a machine-learning-based approximate Bayesian approach, as a faster alternative for parameter estimation in a mechanistic SECIR epidemiological model using COVID-19 ICU occupancy data from Germany. We compared SBI directly with conventional MCMC across multiple epidemic phases and both short and long inference windows. SBI recovered posterior parameter distributions that closely matched MCMC while substantially reducing runtime from approximately 1000 seconds to about one minute on a single GPU for inference using 31-day time series.
Furthermore, SBI scaled much better for inferring parameters from a 201-series, requiring approximately three minutes, compared to 19{,}000 seconds for MCMC.
We additionally quantified posterior agreement using distributional similarity metrics and posterior predictive checks.

Our results show that SBI can provide rapid and scalable Bayesian calibration for epidemiological models, supporting repeated inference and rapid outbreak analysis in time-sensitive public-health settings.

\section*{Introduction}

Mechanistic epidemiological models are widely used to describe infectious-disease dynamics and to support public-health decision making. Representing key biological and epidemiological processes explicitly, such models have been used to estimate transmission dynamics, project epidemic trajectories, assess intervention strategies, and anticipate healthcare-system burden~\cite{Kong2022,10.1093/ije/dyab106,doi:10.1126/sciadv.abd6370,loki2026}. During the COVID-19 pandemic, mechanistic models played a central role in forecasting hospital and intensive care unit (ICU) occupancy and in evaluating the potential effects of non-pharmaceutical interventions, \cite{doi:10.1126/science.abb9789,Flaxman2020}.

Repeated calibration is particularly important~\cite{FUNK201856,10.1093/ije/dyab106,VIBOUD201813}, as transmission patterns, contact behavior, testing practices, healthcare utilization, and intervention measures may change rapidly during an ongoing outbreak and model parameters inferred from one epidemic phase may not be accurate for longer than a few days or weeks. Public-health decisions based on projected hospital or ICU occupancy therefore require inference methods that are capable of updating parameter estimates and projections, including uncertainty, on timescales compatible with operational decision-making. Delays in model calibration risk reducing the value of model-based forecasts, especially when epidemic dynamics change rapidly~\cite{doi:10.1073/pnas.1812594116}.

Bayesian inference provides a principled framework for calibrating epidemiological models because it combines prior knowledge, observational data, and uncertainty into interpretable inferences, thereby enabling forecasts~\cite{MCMC}. In practice, Bayesian calibration is commonly performed using Markov chain Monte Carlo (MCMC) methods, which generate samples from the posterior distribution through repeated evaluations of the likelihood. MCMC-based inference has been applied successfully to a wide range of infectious-disease modeling problems, including influenza, Ebola, and COVID-19 transmission~\cite{Jewell2009,Li2018-si,House2016-oi}.

Despite its strong theoretical guarantees, MCMC is computationally expensive for complex epidemiological models~\cite{Li2018-si,Craiu2014}, sometimes taking months for detailed analyses~\cite{contento2023integrative}. Modern epidemic simulators often involve high-dimensional parameter spaces, nonlinear dynamical systems, stochastic processes, 
or computationally intensive numerical solvers~\cite{doi:10.1073/pnas.1912789117}. As a result, posterior sampling with MCMC, 
which may require thousands to millions of likelihood evaluations, becomes prohibitively slow for repeated analyses. This computational burden limits the applicability of MCMC for repeated near-real-time analyses during rapidly evolving outbreaks.

Simulation-based inference (SBI) has emerged as a promising alternative to MCMC within Bayesian inference for simulator-based models~\cite{doi:10.1073/pnas.1912789117,lueckmann2021benchmarkingsimulationbasedinference}. Instead of relying on repeated likelihood evaluations, SBI learns a probabilistic mapping from simulated data to model parameters using neural density estimation. Neural posterior estimation (NPE), in particular, trains a conditional density estimator on simulated parameter-data pairs and subsequently evaluates the learned posterior approximation on observed data. Once trained, SBI can rapidly generate posterior samples, making SBI attractive for applications that require repeated inference and fast recalibration.

Recent studies have started exploring SBI for epidemiological applications, including transmission-parameter estimation, outbreak reconstruction, and forecasting in stochastic and compartmental epidemic settings~\cite{Wang2024-ap,
Pinotti2025.11.25.690436,wieland2026assessmentsimulationbasedinferencemethods}. However, the use of learned posterior approximations in these scenarios introduces new methodological challenges: the quality of the simulation design, prior specification, simulation budget, neural-network architecture, and training strategy. Moreover, accurate posterior predictive trajectories do not necessarily imply that the inferred posterior distribution over parameters is accurate, because different parameter combinations can produce similar epidemic dynamics. Careful validation against established Bayesian inference methods is therefore essential before SBI can be used as a substitute for conventional Bayesian approaches \cite{MANZANOPATRON2025103580}.

Here, we investigate NPE as a simulation-based alternative to MCMC for Bayesian calibration of a mechanistic (SECIR) model fitted to COVID-19 ICU occupancy data from Germany. We focus on ICU occupancy because of its direct relevance for healthcare-system burden as well as its lower sensitivity to reporting practice and testing intensity. The inference problem includes epidemiological disease progression parameters as well as time-varying effective contact-rate parameters that capture changes in transmission conditions over time.

We compare SBI and MCMC across multiple epidemic phases and inference horizons. First, we evaluate three  operationally relevant 31-day time windows representing distinct phases of the German COVID-19 epidemic in 2020. These shorter time windows reflect an operational setting in which models may need to be recalibrated repeatedly, as new data arrive. Second, we consider a substantially more challenging 201-day reconstruction problem involving multiple transition change points and a higher-dimensional parameter space. We assess agreement between SBI and MCMC using posterior predictive checks as well as quantitative comparisons of marginal posterior distributions based on Wasserstein distance (WD) and Kullback–Leibler divergence (KLD). By combining predictive evaluation with posterior comparison, we aim to determine whether SBI can provide a computationally efficient approximation to MCMC-based Bayesian calibration while preserving the posterior structure relevant for epidemiological interpretation and uncertainty quantification.

\section*{Materials and methods}

\subsection*{The epidemiological model}
\label{sec:model}

\begin{figure}[h!]
    \centering
    \includegraphics[width=\textwidth]{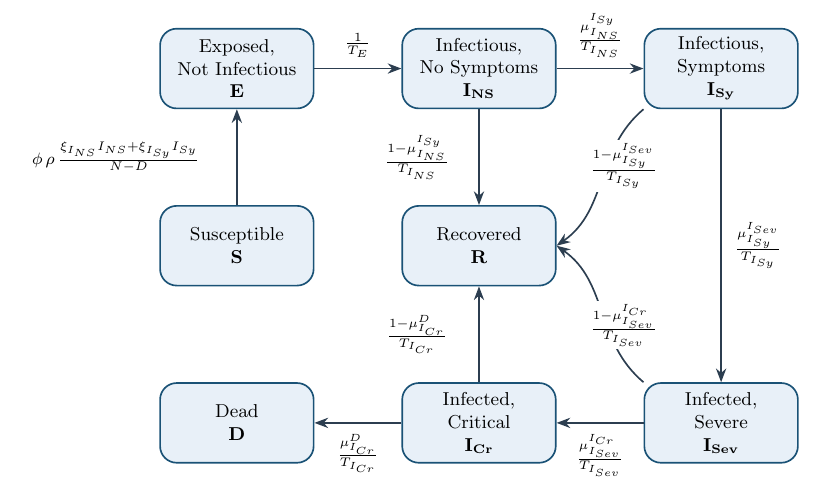}
    \caption{\textbf{Overview of the SECIR model compartment and transition structure.} The model partitions the population into susceptible ($S$), exposed but not yet infectious ($E$), infectious with no symptoms ($I_{NS}$), infectious with symptoms ($I_{Sy}$), infected severe ($I_{Sev}$), infected critical ($I_{Cr}$), recovered ($R$), and dead ($D$) compartments. Arrows indicate transitions between compartments, and edge labels denote the corresponding transition rates. Disease progression is driven by mean transition times $T$ and transition probabilities $\mu$, while infection depends on the effective contact rate, the transmission probability on contact, and the relative infectiousness of non-symptomatic and symptomatic individuals.}
    \label{fig:SECIR}
\end{figure}

We use a mechanistic compartmental model to describe the dynamics  of COVID-19 infections and ICU occupancy in Germany in 2020. The model structure follows the SECIR-formulation introduced in~\cite{kuhn_assessment_2021} and is implemented in the \textit{MEmilio} framework~\cite{bicker2026memilio}. The total population is partitioned into eight epidemiological compartments: susceptible ($S$), exposed ($E$), infectious with no symptoms ($I_{NS}$), infectious with symptoms ($I_{Sy}$), infected severe ($I_{Sev}$), infected critical ($I_{Cr}$), recovered ($R$), and dead ($D$) (see~Fig.~\ref{fig:SECIR}). We use ordinary differential equations (ODEs) to describe the transitions of individuals between these compartments. The model is well suited for modeling the early phase of the COVID-19 pandemic, since represents a largely susceptible population without prior immunity, separates non-symptomatic and symptomatic population, and explicitly tracks disease states, thereby providing important indicators for decision-making. In particular, the $I_{Cr}$ compartment is directly linked to the observed ICU occupancy. 

The ODE system is parameterized by transition times ($T$), transition probabilities ($\mu$), and transmission-related scaling factors ($c, \rho, \xi$). Transition times determine the average duration spent in epidemiological states, while transition probabilities determine the fraction of individuals progressing to the next disease state.
Transition is governed by the effective contact rate $c$ and the transmission probability upon contact $\rho$, and the relative infectiousness of non-symptomatic ($\xi_{I_{NS}}$) and symptomatic ($\xi_{I_{Sy}}$) individuals. Individuals without symptoms either recover or progress to symptomatic infection. Symptomatic individuals then either recover or progress to severe and critical disease states, where they either recover or die.
The full list of the 15 model parameters is given in~\Cref{tab:SECIR}).

To represent changes in transmission conditions over time, the model allows time-dependent reduction to modify the baseline contact rate upon behavioral changes or non-pharmaceutical interventions (NPIs). Let $c_0$ denote the baseline contact rate before the change in transmission conditions. If an intervention is introduced at time $t_1 > t_0$, where $t_0$ is the start of the simulation, the initial contact rate $c_0$ is modified by a reduction factor $r$, where $r$ ranges from 0 (no reduction of contacts) to 1 (complete suppression of contacts). To avoid discontinuities in the ODE system, changes in the contact rate are not applied instantaneously. Instead, the contact rate is smoothed over a short transition interval $\delta \in (0, 1)$ using differentiable interpolation as described in~\cite{kuhn_assessment_2021}. The time-varying contact rate $c(t)$ is thus defined as:

\begin{align}
    \label{eq:damping}
    {c}(t) := \begin{cases} c_0, & t \leq t_1 \\ \widehat{c}(t), & t \in (t_1, t_1 + \delta) \\ (1-r)c_0, & t \geq t_1 + \delta \end{cases}
\end{align}
where $\hat{c}(t)$ denotes the smooth interpolation between $c_0$ and $(1-r)c_0$. The large number of simulations required
for both SBI as well as for the likelihood evaluations for MCMC were performed using the C++ implementation of the SECIR model in the \textit{MEmilio} framework~\cite{bicker2026memilio}.

\begin{table}[h!]
    \centering
    \caption{\textbf{Parameters of the SECIR compartmental model}}
    \label{tab:SECIR}
    \begin{tabular}{|c|c|c|}
        \hline
        Name &  Symbol & Dimension \\
        \hline
        Initial contact rate & $c_0$ & $\mathbb{R}_{\ge 0}$\\
        Reduction factor & $r$ & $[0,1]$\\
        Transmission probability on contact & $\rho$ & $[0,1]$ \\ 
        Relative infectiousness non-symptomatic individuals & $\xi_{I_{NS}}$ & $[0,1]$ \\ 
        Risk infectiousness symptomatic individuals & $\xi_{I_{Sy}}$& $[0,1]$ \\ 
        Population size & $N$  & $\mathbb{R}_{>0}$ \\
        Time exposed & $T_E$  & $\mathbb{R}_{>0}$ \\
        Time infected non-symptomatic & $T_{I_{NS}}$  & $\mathbb{R}_{>0}$ \\
        Time infected symptomatic & $T_{I_{Sy}}$  & $\mathbb{R}_{>0}$\\
        Time infected severe & $T_{I_{Sev}}$ & $\mathbb{R}_{>0}$ \\
        Time infected critical & $T_{I_{Cr}}$ & $\mathbb{R}_{>0}$ \\
        Transition prob. from $I_{NS}$ to $I_{Sy}$ & $\mu^{I_{Sy}}_{I_{NS}}$ & $[0,1]$ \\
        Transition prob. from $I_{Sy}$ to $I_{Sev}$ & $\mu^{I_{Sev}}_{I_{Sy}}$ & $[0,1]$ \\
        Transition prob. from $I_{Sev}$ to $I_{Cr}$ & $\mu^{I_{Cr}}_{I_{Sev}}$ & $[0,1]$ \\
        Transition prob. from $I_{Cr}$ to $D$ & $\mu^{D}_{I_{Cr}}$ & $[0,1]$ \\
        \hline
    \end{tabular}
\end{table}

\subsection*{Data}

The reported ICU occupancy of COVID-19 patients~\cite{robert_koch_institut_2025_ICU} in Germany from April 24 to December 10 2020 serves as the empirical basis for comparing SBI and MCMC powered inference. We use the ICU occupancy rather than reported daily case count because case counts are more strongly affected by changes in testing intensity, reporting delays, and under-reporting~\cite{Streeck2020, Gornyk2021}. Furthermore, ICU occupancy is also directly relevant for healthcare-system burden and public-health decision making. Let $x^{obs}_{1:231}=(x^{obs}_1,\dots,x_{231}^{obs})$ denote the observed ICU occupancy time series over this period. For our inference experiments below, we consider different time windows $x^{obs}_{t_0:t_{end}}$, where $t_0$ is the respective starting time and $t_{end}$ is the end time.

\subsection*{Bayesian parametrization}


\subsubsection*{Estimated parameters}

The Bayesian model was designed to infer epidemiological parameters together with a time-varying effective contact rate.
Specifically, the Bayesian model estimates the five transmission-time parameters 
$T_E$, $T_{I_{NS}}$, $T_{I_{Sy}}$, $T_{I_{Sev}}$, and $T_{I_{Cr}}$ as well as the 
relative infectiousness of symptomatic individuals $\xi_{I_{Sy}}$.
In addition, the Bayesian model infers changes in the time-varying contact rate on a grid. Let $\Delta$ be the minimum number of days for which a contact-rate level is valid, then for a time series starting at day $t_0$ and ending after day $t_{end}$ the number of intervals on the grid is $K=\left\lfloor{\tfrac{t_{end}-t_0+1}{\Delta}}\right\rfloor$.
For $\tau_k=t_0+k\Delta, k=0,\dots,K$ the unsmoothed effective time-varying contact rate $\beta_{\mathrm{pc}}(t)$ is given by the piecewise-constant function
\begin{equation}
    \label{eq:effective_dampening}
    {\beta_{\mathrm{pc}}}(t) := \rho \cdot c_0 \cdot (1-r_k)
    \, \text{for}\, t\in[\tau_{k-1}, \tau_{k})
    ,\qquad k=1,\dots,K,
\end{equation} 
where the $r_k$ are reduction factors for each of the $K$ intervals. The effective contact rate used in the ODE system is obtained by smoothing transitions between consecutive levels of $\beta_{\mathrm{pc}}(t)$ over a short transition as described above (Eq.~\eqref{eq:damping}).
As before, larger values for $r_k$ correspond to stronger reduction in the effective contact rate (1 representing reduction to 0 contacts and 0 representing no reduction of contacts). To avoid non-identifiability of the baseline contact rate and the transition probability on contact, we fix $c_0=1$ and $\rho=1$, meaning that for each interval on the grid the effective contact rate is controlled by a single parameter $r_k$.

The full parameter vector of the Bayesian model is given by $\theta$.
\begin{equation}
\theta=(T_E, T_{I_{NS}}, T_{I_{Sy}}, T_{I_{Sev}},T_{I_{Cr}}, \xi_{I_{Sy}},
r_1,\dots,r_K)
\end{equation}

\subsubsection*{Model Initialization}

To initialize the epidemiological compartments for inference starting at time $t_0$ we use the initialization procedure described in~\cite{zunker_novel_2024}. This procedure combines reported case data, ICU occupancy data and population data. Reported confirmed cases~\cite{robert_koch_institut_2025_17627234} are mapped to the exposed, (a)symptomatic infectious, severe, critical, recovered and deceased compartments by shifting the case time series according to assumed transition times and weighting the resulting contributions with the corresponding disease progression probabilities. 
ICU occupancy was obtained from the DIVI intensive care registry~\cite{robert_koch_institut_2025_ICU}. 
Population data~\cite{regionaldatenbank_deutschland_fortschreibung_2022} is used to set the susceptible compartment as the remaining population after all other compartments have been initialized.

\subsubsection*{Prior distributions}

Because ICU occupancy provides information on only one 
of the eight epidemiological compartments, prior information is required to constrain weakly identifiable disease-progression and transmission parameters. We therefore used literature-informed priors for the epidemiological parameters and
bounded uniform priors for the time-varying reduction factors~\cite{Schilling2020Krankheitsschwere,Byambasuren2020Asymptomatic,Dhouib2021Incubation}. The prior distributions are summarized in~\Cref{tab:prior}, while the priors on the reduction factors for 201-day window are summarized in \nameref{S1_Table}, Spec B.

\begin{table}[h!]
    \centering
    \caption{\textbf{Priors for the Bayesian model.}
    The median and 95\% intervals cover reasonable values for the SECIR model.
    }
    \label{tab:prior}
    \begin{tabular}{|c|c|}
        \hline
        Prior & Description \\
        \hline
        $T_E \sim \operatorname{Uniform(2.67, 4.0)}$  & 
        median 3.36; 95\% interval [2.7, 3.97]
        \\
        \hline
        $T_{I_{NS}} \sim \operatorname{LogNormal}(\ln(10),0.2)$  & 
        median 10; 95\% interval [6.76, 14.80] 
        \\
        \hline
        
        $T_{I_{Sy}} \sim \operatorname{LogNormal}(\ln(10),0.2)$  & 
        median 10; 95\% interval [6.76, 14.80] 
        \\
        
        $T_{I_{Sev}}
        \sim \operatorname{LogNormal}(\ln(10),0.2)$  & 
        median 10; 95\% interval [6.76, 14.80] 
        \\
        \hline
        
        $T_{I_{Cr}}
        \sim \operatorname{LogNormal}(\ln(10),0.2)$  & 
        median 10; 95\% interval [6.76, 14.80] 
        \\
        \hline
        
        $\xi_{I_{Sy}} \sim \operatorname{Uniform}(0.01, 0.9)$ & 
        median 0.46; 95\% interval [0.03, 0.88] 
        \\
        \hline
        $r_k \overset{ind}{\sim} \operatorname{Uniform}(0,1)$ & 
        median 0.5; 95\% interval [0.025, 0.975]
        \\
        \hline
        
    \end{tabular}
\end{table}

\subsubsection*{Prior and posterior predictive checks}

Prior and posterior predictive checks were used to assess the consistency of the Bayesian model before and after conditioning on the observed ICU occupancy data.

For prior predictive checks, 1000 parameter samples  were drawn from the prior distribution and propagated through the SECIR model to generate prior predictive trajectories. These trajectories were compared with the observed ICU occupancy data to evaluate whether the prior distributions generated epidemiologically plausible epidemic dynamics.

For posterior predictive evaluation, 1000 posterior samples were drawn from the inferred posterior distribution. Each sampled parameter vector was propagated through the SECIR model to generate one simulated ICU occupancy trajectory. The resulting ensemble of simulated trajectories was used to approximate the posterior predictive distribution. For each time point, we computed the posterior predictive mean and the pointwise \SI{95}{\percent} credible interval. As a quantitative measure of predictive performance, we calculated the root mean squared error (RMSE) between the trajectory generated from the posterior predictive mean trajectory and the observed ICU occupancy data.

Prior predictive checks were used to assess the suitability of the prior specification, whereas posterior predictive checks were used to evaluate whether the inferred posterior distributions reproduced the observed epidemic dynamics.

\subsection*{Markov chain Monte Carlo reference inference}

To compute baseline inferences for comparison we 
use Markov chain Monte Carlo (MCMC).
MCMC is especially suited to create such baseline inferences as it has strong theoretical
guarantees that samples will be asymptotically distributed
according the targeted posterior distribution.

\subsubsection*{Observation model and likelihood}

For MCMC inference, we used an explicit likelihood function relating simulated ICU occupancy to observed ICU occupancy.
Following~\cite{dehning2023impact}, we assumed Student's t-distributed observation errors with $\nu=4$ degrees of freedom. Student's t-distribution is widely used to construct likelihoods for noisy observations as it's less sensitive to outliers. For an observation window from $t_0$ to $T$, the likelihood was
\begin{equation}
    p(x^{obs}_{t_0:T}|\theta,t_{0:T})=
        \prod_{i=t_0}^{T} p(x^{obs}_i|\theta, t_i)=
        \prod_{i=t_0}^{T} \text{Student-}t_{\nu}(x^{obs}_i|x^{sim}(\theta, t_i), \sigma_i)
\end{equation}
where $x^{sim}(\theta, t_i)$ denotes the simulated ICU occupancy at day $t_i$ and the observation noise modeled as 
$\sigma_i=\sqrt{|x^{sim}(\theta, t_i)|}$. 

\subsubsection*{MCMC reference inference}

We used MCMC as the reference Bayesian inference method against which SBI was compared. MCMC is suitable for this purpose because, under standard regularity conditions and sufficient convergence, it provides samples from the targeted posterior distribution.
Posterior sampling was performed using the differential-evolution Metropolis-Z algorithm (DE-MZ)~\cite{terBraakVrugt2008}. DE-MZ is an adaptive, gradient-free MCMC method that uses the history of the MCMC chains to adapt proposals/samples to the scale and correlation structure of the posterior during sampling.
Both DE-MZ and SBI do not require the evaluation of the likelihood gradient, which is computationally expensive for the SECIR ODE system. 

We used the DE-MZ implementation provided by \pymc{} v5.0.0~\cite{pymc2023}.
Tuning is implemented by \pymc{}. 
As a rule, for $n$ samples we want to draw, we sample an additional $n$ samples beforehand for tuning and discard them once tuning is done.
Convergence was assessed using rank-normalized $\hat{R}$ diagnostic~\cite{vehtari2021rank} and check that
the value is close to 1. 

\subsection*{Simulation Based Inference}

Simulation-Based Inference (SBI) leverages Artificial Intelligence to construct an approximate Bayesian framework. Here, we use the Neural Posterior Estimation (NPE) method \cite{lueckmann2019}, which approximates the posterior distribution of the model parameters. 

\subsubsection*{Amortized Neural Posterior Estimation.}

We used simulation-based inference with Neural Posterior Estimation (NPE) to approximate the posterior distribution over model parameters. NPE learns a conditional density estimator from simulated parameter--data pairs and subsequently generates approximate posterior samples for observed data without requiring repeated likelihood evaluations.

The objective is to approximate the Bayesian posterior distribution

\begin{equation}
p(\theta \mid X) \propto L(X \mid \theta)\pi(\theta),
\end{equation}

where $L(\cdot)$ denotes the likelihood function and $\pi(\cdot)$ the prior distribution. Rather than evaluating the likelihood explicitly, NPE trains a neural density estimator using simulations generated from the prior predictive distribution.

Training data were generated by sampling parameter vectors $\theta_j \sim \pi(\theta)$ and propagating them through the SECIR simulator to obtain synthetic ICU occupancy trajectories $\hat X_j = f(\theta_j)$, where $f$ denotes the epidemiological simulator. This produced a set of simulated parameter--data pairs $(\theta_j,\hat X_j)$ used to train the neural density estimator.

The network learns a conditional posterior approximation $\tilde p(\theta \mid \hat X)$, where the simulated trajectory serves as conditioning information and the distribution over model parameters is represented directly. In this way, the network learns a mapping from simulated epidemic trajectories to posterior distributions over parameters.

\begin{figure}
\includegraphics[width=\textwidth]{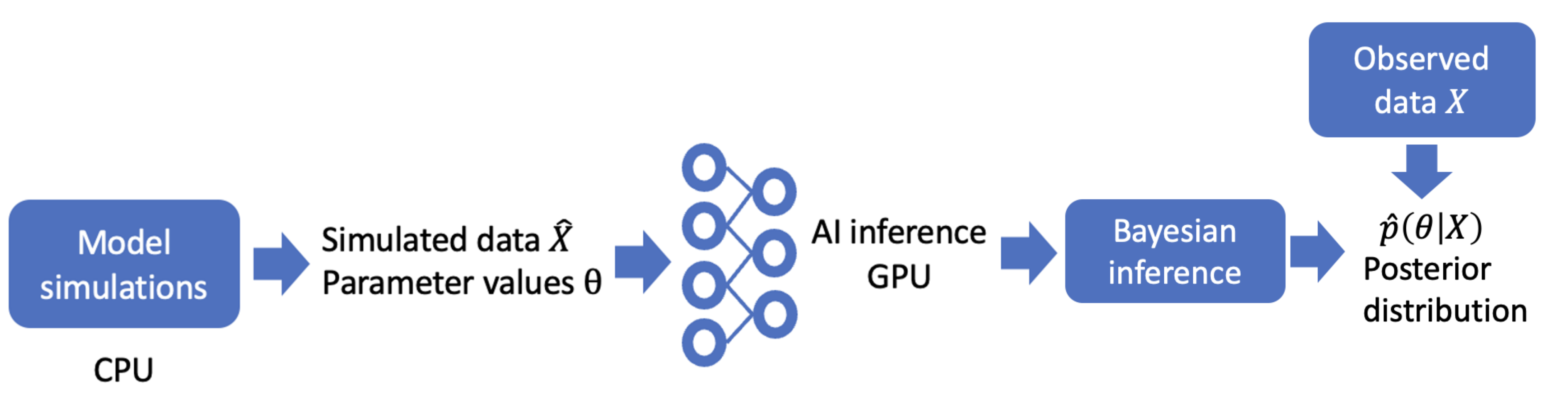}\caption{\textbf{Workflow of amortized neural posterior estimation (NPE).} Parameters $\theta$ are sampled from the prior distribution and propagated through the SECIR epidemiological model to generate simulated ICU occupancy trajectories $\hat{X}$. The resulting parameter–trajectory pairs $(\theta,\hat{X})$ are used to train a neural density estimator that approximates the conditional posterior distribution $p(\theta|X)$. After training, the observed ICU occupancy data $X$ are provided as conditioning information, and samples are drawn from the learned posterior approximation $\hat{p}(\theta|X)$. Because the posterior estimator is trained once and subsequently reused for inference, the approach is amortized.}\label{fig:sbi_schema}
\end{figure}

The approach is amortized because the posterior estimator is trained once using a large collection of simulated parameter--data pairs and can subsequently be applied to any new observation generated by the same simulator and prior distribution. After training, the observed ICU occupancy data $X$ are provided as conditioning information and parameter samples are drawn from the learned posterior approximation $\tilde p(\theta \mid X)$. A schematic overview of the procedure is shown in Fig.~\ref{fig:sbi_schema}.

\subsubsection*{Neural Density Estimator and Trajectory Embedding}\label{sec:MAF}

The posterior distribution was approximated using a Masked Autoregressive Flow (MAF) \cite{papamakarios2018maf}, a normalizing-flow density estimator commonly used for neural posterior estimation. In the neural posterior estimation framework, MAF models the conditional posterior distribution autoregressively as
\begin{equation}
p(\theta \mid \hat X)=
\prod_{i=1}^{d}
p(\theta_i \mid \theta_{1}, \hat X),
\end{equation}
where $\hat X$ denotes the simulated ICU occupancy trajectory used as conditioning information and $d$ is the dimension of the parameter vector. Additional mathematical details on normalizing flows, MAF, and the underlying Masked Autoencoder for Density Estimation (MADE) architecture are provided in the Supplementary Information, \nameref{S1_Text}.

Because the simulated ICU occupancy trajectories are time series, we employed a Convolutional Neural Network (CNN) embedding to extract a lower-dimensional representation before conditioning the density estimator. Rather than providing the full trajectory directly to the MAF, the embedding network maps the simulated trajectory $\hat X$ to a learned feature representation $h(\hat X)$, which is subsequently used as conditioning information. CNNs are well suited for extracting informative local and multiscale features from one-dimensional temporal data \cite{726791}. The final architecture consisted of a four-layer CNN embedding network followed by a MAF density estimator with ten hidden features and two flow transformations.


Training was performed using the Sequential Neural Posterior Estimation (SNPE-C) objective \cite{greenberg2019,lueckmann2019}. The loss combines a contrastive (atomic) term with a maximum-likelihood regularization term,
\begin{equation}
\mathcal{L}_{\mathrm{atomic}}
+
\mathcal{L}_{\mathrm{MLE}},
\end{equation}
where the regularization term reduces density leakage outside the support of the prior distribution. Additional details on the SNPE-C objective are provided in the Supplementary Information (\nameref{S2_Text}).

\subsubsection*{SBI implementation}

Parameter inference was performed using the \texttt{sbi} Python library (version 0.22.0; \cite{Tejero-Cantero2020}). Within the \texttt{memilio} framework, we developed a dedicated SBI wrapper compatible with the \texttt{simulate\_for\_sbi} interface and parameter priors specified through \texttt{MultipleIndependent} distributions.

The simulated ICU occupancy trajectories were embedded using a convolutional neural network (CNN) constructed via the \texttt{CNNEmbedding} function. To assess the effect of the embedding architecture, we evaluated CNNs with between two and four convolutional layers and output channel sizes ranging from 6 to 12 channels per layer. Candidate architectures included channel configurations such as $[6,12,12]$ and $[6,12,12,12]$. The resulting embedding was used as conditioning information for a MAF density estimator with ten hidden features and two flow transformations. Posterior inference was performed using the SNPE-C algorithm.

To investigate the sensitivity of SBI to training hyperparameters, we evaluated simulation budgets of \num{20000}, \num{50000}, and \num{100000} forward simulations in combination with batch sizes of 450, 900, 1800, 3600, 7200, 14400, 22500, 45000, and 90000. 

To improve training performance for large simulation datasets, we implemented a custom PyTorch \texttt{Dataset} that uses vectorized tensor indexing and avoids repeated preprocessing while preserving the standard SNPE-C training and inference workflow. Training was performed on a GPU using the custom dataloader.

After training, the observed ICU occupancy trajectory was supplied as conditioning information to the learned posterior estimator, enabling sampling from the approximate posterior distribution over model parameters.

\subsection*{Comparison of posterior distributions}
To quantify agreement between posterior distributions obtained via simulation-based inference (SBI) and Markov chain Monte Carlo (MCMC), we compared the marginal posterior distributions of each inferred parameter using the 1-Wasserstein distance and the Kullback--Leibler (KL) divergence.

For a given scalar parameter $\theta_j$, let $p(\theta_j)$ denote the marginal posterior density inferred by SBI and $q(\theta_j)$ the corresponding marginal posterior density inferred by MCMC. Since both posteriors are represented by samples, continuous density estimates $\hat p(\theta_j)$ and $\hat q(\theta_j)$ were obtained using Gaussian kernel density estimation (KDE). The densities were evaluated on a uniform grid spanning the combined support of both sample sets and normalized numerically.

\paragraph{Wasserstein distance.}
The first-order Wasserstein distance (Earth Mover's Distance) between the two marginal posterior distributions was computed as
\begin{equation}
W_1(p,q)=\inf_{\gamma \in \Pi(p,q)}
\mathbb{E}_{(x,y)\sim\gamma}
\left[|x-y|\right],
\end{equation}
where $\Pi(p,q)$ denotes the set of all joint distributions with marginals $p$ and $q$. In practice, this quantity was computed directly from the posterior samples using a standard empirical estimator.

\paragraph{Kullback--Leibler divergence.}
The KL divergence from the SBI posterior to the MCMC posterior was defined as
\begin{equation}
D_{\mathrm{KL}}(p(\theta_j)||q(\theta_j))=\int
\hat p(\theta_j)
\log
\frac{\hat p(\theta_j)}
{\hat q(\theta_j)} d\theta_j .
\end{equation}

The integral was approximated numerically on the discretized grid,
\begin{equation}
D_{\mathrm{KL}}(p(\theta_j)||q(\theta_j))
\approx
\sum_{k=1}^{K}
\hat p(\theta_{j,k})
\log
\frac{\hat p(\theta_{j,k})}
{\hat q(\theta_{j,k})}
\Delta\theta ,
\end{equation}
where $\Delta\theta$ denotes the grid spacing. To improve numerical stability, density values were clipped to a small positive constant before evaluating the logarithm.
\paragraph{Symmetric KL divergence.}
Because the KL divergence is asymmetric, we additionally computed the symmetric KL divergence,
\begin{equation}
D_{\mathrm{sym}}\!\left(p(\theta_j),q(\theta_j)\right)
=
\frac12
\left[
D_{\mathrm{KL}}\!\left(p(\theta_j)\|q(\theta_j)\right)
+
D_{\mathrm{KL}}\!\left(q(\theta_j)\|p(\theta_j)\right)
\right].
\end{equation}
All metrics were computed independently for each inferred parameter. For each inference configuration, we report the mean and standard deviation of the resulting distances across 16 independent SBI runs.
\section*{Results}

\subsection*{Experimental setups}


We compared SBI with MCMC of the SECIR model to COVID-19 ICU occupancy data from Germany. Inference was performed for three 31-day inference windows and one 201-day inference window representing distinct epidemic phases, including a declining post-first-wave phase of June 2020 with low incidence summer period and two increasing autumn 2020 phases, started at offsets of 0, 160, and and 200 days after April 24, 2020. The extended 201-day window was used to assess performance in a long and higher-dimensional reconstruction problem involving multiple changes in transmission conditions.

For the 31-day windows, the selected SBI configuration used \num{50000} simulations and a batch size of \num{14400}. For the 201-day inference problem, which involved a higher-dimensional parameter space and multiple change points, the selected configuration used \num{100000} simulations and a batch size of \num{1800}. Each SBI configuration was repeated 16 times. Posterior predictive performance, posterior agreement with MCMC, and runtime are summarized in Table~\ref{tab:rmse}.

\subsection*{Prior predictive performance}

Prior predictive checks were particularly important for the 201-day inference problem because the larger number of change-point parameters substantially increased the flexibility of the model. While most epidemiological parameters were constrained by literature-informed priors, broad prior bounds on the change-point parameters produced highly dispersed trajectories, including implausible epidemic dynamics and partially non-identifiable parameter configurations (Fig.~\nameref{S1_Fig}, left). As a result, a substantial proportion of simulations occupied regions that were inconsistent with the observed epidemic trajectory.

The sensitivity of the model to change-point parameters is amplified over the longer 201-day reconstruction window, where relatively small parameter changes can lead to markedly different epidemic trajectories. For SBI, priors that support implausible epidemic dynamics impose an avoidable burden on training. While principled methods are being developed to meaningfully constrain priors~\cite{Drovandi2025}, here we adopted tighter prior bounds for the change-point parameters in both the MCMC and SBI analyses. The revised specification retained sufficient variability to capture a broad range of plausible epidemic scenarios while excluding unrealistic dynamics.

Under the final prior specification, the observed ICU occupancy trajectory was well covered by the prior predictive envelope (Fig.~\nameref{S1_Fig}, right), indicating that the priors were neither overly restrictive nor unrealistically diffuse.

\subsection*{Posterior predictive performance}

\begin{figure}[htbp]
\centering

\includegraphics[width=\textwidth]{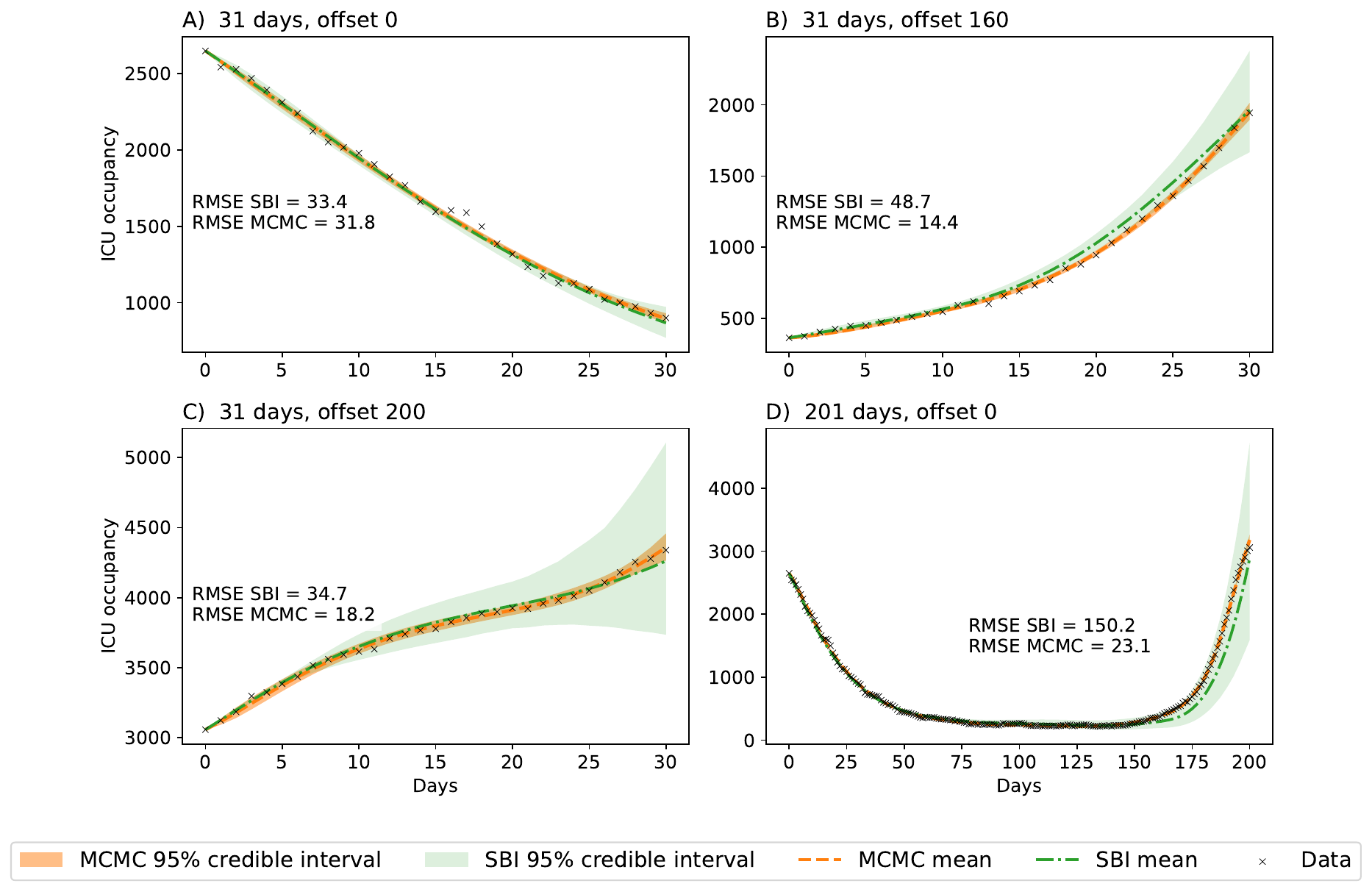}

\caption{{\bf Posterior predictive checks for the inferred SBI posterior distributions across the analyzed inference windows.} For each setting, ICU occupancy trajectories were simulated using parameter samples drawn from the inferred posterior distribution. The solid line shows the posterior predictive mean trajectory, the shaded region indicates the 95\% pointwise credible interval, and the black line corresponds to the observed ICU occupancy data. The panels include the three 31-day inference windows starting at offsets 0 (upper left), 160 (upper right), and 200 (lower left) days after April 24, 2020, as well as the extended 201-day (lower right) inference window. Across all settings, the posterior predictive distributions capture the dominant temporal dynamics of the observed epidemic trajectory.}
\label{fig:four_panel}
\end{figure}

Figure~\ref{fig:four_panel} shows representative posterior predictive checks for the selected inference settings. As a quantitative summary, Table~\ref{tab:rmse} reports the mean RMSE and standard deviation across 16 independently trained SBI models.

\begin{table}[ht]
\centering
\scriptsize
\setlength{\tabcolsep}{4pt} 

\resizebox{\textwidth}{!}{%
\begin{tabular}{|l|lllll|}
\hline
                    & Wasserstein & KL             & KL sym         & RMSE            & Runtime          \\
                    \hline
offset 0, 31 days   &             &                &                &                 &                  \\
\hline
900                 & 0.98 (1.14) & 17.18 (42.13)  & 9.35 (21.37)   & 153.52 (384.73) & 175.19 (102.72)  \\
14400               & 0.75 (0.51) & 17.92 (24.98)  & 9.58 (13.2)    & 66.41 (26.16)   & 71.47 (19.22)    \\
45000               & 0.63 (0.37) & 17.3 (24.38)   & 8.92 (12.36)   & 70.43 (62.9)    & 63.13 (13.32)    \\ \hline
offset 160, 31 days &             &                &                &                 &                  \\ \hline
900                 & 3.38 (2.5)  & 189.49 (69.15) & 134.67 (73.38) & 115.07 (23.34)  & 166.1 (27.65)    \\
14400               & 2.37 (1.43) & 138.17 (53.13) & 75.79 (31.35)  & 85.88 (54.78)   & 60.83 (3.09)     \\
45000               & 2.47 (1.34) & 146.1 (54.44)  & 78.8 (31.63)   & 137.86 (103.24) & 61.01 (6.67)     \\
\hline
offset 200, 31 days &             &                &                &                 &                  \\ \hline
900                 & 1.90 (0.49) & 87.6 (5.59)    & 58.46 (18.64)  & 68.72 (37.52)   & 161.11 (27.65)   \\
14400               & 1.89 (0.45) & 85.44 (12.33)  & 46.44 (9.58)   & 98.79 (66.22)   & 60.07 (3.97)     \\
45000               & 1.98 (0.34) & 84.6 (13.6)    & 47.06 (10.82)  & 144.51 (140.1)  & 59.68 (5.02)     \\
\hline
offset 0, 201 days  &             &                &                &                 &                  \\ \hline
450                 & 0.50 (0.28) & 61.8 (56.6)    & 39.01 (42.59)  & 323.43 (126.72) & 874.51 (1000.34) \\
1800                & 0.35 (0.35) & 141.69 (47.11) & 72.14 (24.67)  & 325.72 (135.78) & 157.08 (42.28)   \\
22500               & 0.46 (0.18) & 132.25 (44.12) & 66.98 (22.65)  & 397.93 (129.21) & 86.55 (9.23)     \\
\hline
\end{tabular}%
}

\caption{{\bf Comparison of posterior agreement, posterior predictive performance, and runtime across inference configurations.} For each inference window and batch size, we report the mean and standard deviation across 16 repeated runs of the 1-Wasserstein distance, Kullback-Leibler divergence, symmetric Kullback-Leibler divergence between SBI and MCMC posterior distributions, posterior predictive RMSE, and total SBI runtime in seconds. The selected configurations used for the main analyses were 50{,}000 simulations with batch size 14{,}400 for the 31-day inference windows and 100{,}000 simulations with batch size 1{,}800 for the 201-day inference window. All SBI experiments were performed on a single NVIDIA A100 GPU.}
\label{tab:rmse}
\end{table}

Across the three 31-day windows, SBI reproduced the observed ICU trajectories well. Across the selected short-window configurations, the offset 0 window achieved the lowest RMSE of $66.41 \pm 26.16$. This window corresponds to the declining phase after the first epidemic wave. The offset 160 and offset 200 windows yielded RMSE values of $85.88 \pm 54.78$ and $98.79 \pm 66.22$, respectively. These later windows involved more rapidly changing epidemic dynamics and showed broader posterior predictive uncertainty, but the posterior predictive trajectories still followed the observed ICU occupancy patterns. Compared with the MCMC reference posterior predictive distributions, which achieved RMSE values of 31.8, 14.4, and 18.2 for the offset 0, 160, and 200 windows, respectively, SBI generally produced broader credible intervals while retaining similar overall trajectory shapes and temporal trends.

The 201-day inference problem was substantially more challenging because the model had to reconstruct both the decline after the first wave and the subsequent increase during the second wave within a single inference window. The selected 201-day SBI configuration yielded an RMSE of $325.72 \pm 135.78$. This error was larger than for the 31-day windows, as expected from the longer reconstruction horizon, the higher-dimensional parameter space, and the need to infer multiple change-point parameters simultaneously. Nevertheless, the posterior predictive trajectories captured the dominant temporal structure of the observed ICU data, including the prolonged low-incidence period and the sharp autumn increase in ICU occupancy. The corresponding MCMC posterior predictive distribution achieved an RMSE of 23.1 and produced narrower credible intervals, although both approaches reproduced the major epidemic phases observed in the ICU occupancy data.

\subsection*{Agreement between SBI and MCMC posteriors}

\begin{figure}[htbp]
\centering

\includegraphics[width=\textwidth,height=0.82\textheight,keepaspectratio]{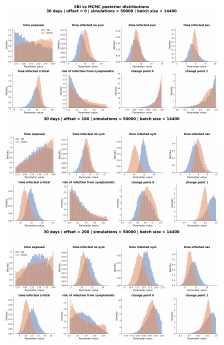}

\caption{{\bf Comparison of marginal posterior distributions inferred using simulation-based inference (SBI, blue) and Markov chain Monte Carlo (MCMC, orange).} The panels show posterior distributions for epidemiological parameters and change-point parameters inferred from ICU occupancy data for the 31-day windows starting at offsets 0 (upper panel), 160 (middle panel), and 200 (lower panel) days after April 24, 2020. Across all settings, SBI recovered posterior modes and overall parameter trends broadly consistent with the MCMC reference posteriors.}
\label{fig:comparison}
\end{figure}

We next compared the posterior distributions inferred by SBI with the corresponding MCMC reference posteriors. Agreement was assessed visually using marginal posterior distributions, see Fig. \ref{fig:comparison}, \nameref{S2_Fig} and quantitatively using the first Wasserstein distance, KL divergence, and symmetric KL divergence, Tab \ref{tab:rmse}.

For the 31-day inference windows, SBI recovered marginal posterior distributions that were broadly consistent with the MCMC reference posteriors (Fig.~\ref{fig:comparison}). The strongest agreement was observed for the offset 0 window (top panel). In this setting, the selected SBI configuration achieved an average first-order Wasserstein distance of $0.75 \pm 0.51$ and a symmetric KL divergence of $9.58 \pm 13.2$. The posterior modes inferred by SBI and MCMC were closely aligned for most epidemiological and reduction factor parameters.

Posterior agreement remained good for the offset 160 and offset 200 windows, although discrepancies increased for some parameters. In these later windows, SBI generally produced broader marginal posterior distributions than MCMC. This broadening was most visible for selected disease-duration and reduction factors. For offset 160, the selected SBI configuration yielded a Wasserstein distance of $2.37 \pm 1.43$ and a symmetric KL divergence of $75.79 \pm 31.35$. For offset 200, the corresponding values were $1.89 \pm 0.45$ and $46.44 \pm 9.58$. Despite these larger divergences, the dominant posterior regions remained aligned between SBI and MCMC.

The 201-day inference problem involved a substantially larger parameter space and showed broader posterior uncertainty than the 31-day analyses, \nameref{S2_Fig}. Nevertheless, SBI preserved the main posterior structure observed in the MCMC reference. For the selected 201-day configuration, the Wasserstein distance was $0.35 \pm 0.35$ and the symmetric KL divergence was $72.14 \pm 24.67$. Several reduction factor parameters showed substantial overlap between SBI and MCMC, whereas larger discrepancies were observed for parameters near the end of the observation window. These late change points were less strongly constrained by the data and therefore showed broader posterior support under SBI.

Overall, SBI recovered posterior distributions that were consistent with the MCMC reference posteriors across all inference settings. While the SBI marginals were generally smoother and somewhat broader than the corresponding MCMC distributions, the dominant posterior regions and parameter trends were preserved.

\subsection*{Effect of simulation budget and batch size}


We evaluated the effect of simulation budget and batch size on SBI performance using posterior predictive RMSE, posterior distance metrics, runtime, and variability across repeated runs (Table~\ref{tab:rmse}, \nameref{S5_Table}).

For the 31-day inference windows, configurations trained with \num{20000} simulations generally showed higher RMSE and greater variability across repeated runs. Increasing the simulation budget to \num{100000} did not consistently improve posterior predictive performance or agreement with the MCMC reference posteriors. Across the three 31-day windows, \num{50000} simulations provided stable posterior recovery while maintaining substantially lower computational cost. Among the tested batch sizes, a batch size of \num{14400} yielded consistently low RMSE, good agreement with MCMC, and stable runtimes, and was therefore selected for the subsequent analyses.

For the 201-day inference problem, the optimal configuration differed from that observed for the 31-day windows. Because this setting involved 21 inferred parameters and multiple reduction factors, posterior recovery was more sensitive to the training setup. The configuration with \num{100000} simulations and batch size \num{1800} achieved the most stable posterior agreement with MCMC across repeated runs while maintaining competitive posterior predictive performance. Larger batch sizes did not improve posterior agreement or RMSE.

Across experiments, very small batch sizes increased both runtime and variability across repeated runs. For the 31-day inference windows, the smallest tested batch size (\num{900}) required approximately 160--175 seconds depending on the inference window, whereas larger batch sizes yielded substantially lower and more stable runtimes (Table~\ref{tab:rmse}). In several runs, the increased runtime was partly attributable to inefficient posterior sampling from the trained neural posterior, where low acceptance rates increased the time required to generate posterior samples. Very large batch sizes did not consistently improve posterior agreement with MCMC or posterior predictive accuracy. Overall, the optimal hyperparameter configuration differed between the 31-day and 201-day inference settings, indicating that performance depended jointly on simulation budget, batch size, and inference dimensionality.


\subsection*{Consistency across inference windows}

We additionally compared inferred parameters across the short- and long-window analyses to assess whether consistent posterior structures were recovered across different epidemic phases. For the epidemiological parameters that are not explicitly time dependent in the model, the exposed period and the durations of severe and critical infection showed substantial overlap across the three 31-day windows, with similar posterior regions inferred by both SBI and MCMC. In contrast, the symptomatic infection duration shifted toward larger values in the offset 160 and offset 200 windows compared with offset 0, and this pattern was observed for both inference methods. The relative infectiousness of symptomatic individuals remained broadly consistent between offset 0 and offset 160, but shifted toward larger values in the offset 200 window.

The 201-day inference provided an additional consistency check because it spans the same calendar period covered by the shorter 31-day windows within a single joint inference problem. Early reduction factor parameters in the 201-day posterior exhibited similar contact-reduction behavior to the offset 0 inference window, while later reduction factors reflected the stronger transmission dynamics inferred in the offset 160 window. Although the posterior distributions were generally broader in the 201-day setting, reflecting the larger number of inferred reduction factors and stronger parameter correlations, the dominant transmission patterns remained consistent between the short- and long-window analyses.

\subsection*{Runtime comparison}

Details for all MCMC runs are reported in \nameref{S2_Table} and all $\hat{R}$ values are
summarized in Table \nameref{S3_Table}.
SBI substantially reduced runtime compared with the MCMC reference inference. For the 31-day windows, MCMC required approximately \SI{1077}{\second} per inference run, achieving a maximum rank-normalized $\hat R$ value of 1.0119 across all inferred parameters. All MCMC runs were performed on an AMD EPYC 9334 32-Core Processor, whereas SBI experiments were performed on a single NVIDIA A100 GPU. In contrast, the selected SBI configuration required $71.47 \pm 19.22$ seconds for the offset 0 window, $60.83 \pm 3.09$ seconds for the offset 160 window, and $60.07 \pm 3.97$ seconds for the offset 200 window.

The runtime advantage was also observed for the 201-day inference problem. The MCMC reference inference required \num{19102} seconds, whereas the selected SBI configuration required $157.08 \pm 42.28$ seconds. Thus, even in the higher-dimensional long-window setting, SBI remained computationally tractable and substantially faster than MCMC. Despite the high run-time, MCMC failed to achieve low rank-normalized $\hat{R}$ (maximum $\hat{R}=1.605$) for the parameters.

Overall, SBI reduced inference time by approximately one order of magnitude for the 31-day windows and more than two orders of magnitude for the 201-day inference problem while maintaining comparable posterior predictive performance and broad agreement with the MCMC posterior distributions.


\section*{Discussion}


In this study, we evaluated SBI with NPE as a computationally efficient alternative to MCMC for Bayesian calibration of a mechanistic SECIR epidemiological model fitted to COVID-19 ICU occupancy data from Germany. Across multiple epidemic phases and inference horizons, SBI recovered posterior distributions that were broadly consistent with the MCMC reference. By combining posterior predictive checks with quantitative comparisons of posterior distributions, we found that SBI preserved the dominant posterior structure while substantially reducing computational cost. 
These results show that NPE is capable to provide a practical approximation to MCMC-based Bayesian calibration for the SECIR ICU occupancy model considered here.

A central finding of this work is that SBI maintained good posterior predictive performance across qualitatively different epidemic regimes. The three 31-day windows included both declining and rapidly increasing ICU occupancy trajectories, yet the inferred SBI posteriors generated posterior predictive simulations that closely followed the observed data in all cases. This is particularly relevant for operational epidemiological modelling, where models must be recalibrated repeatedly as transmission dynamics change. 
The 201-day inference problem was substantially more demanding because the model had to jointly capture the decline following the first epidemic wave, the low incidence summer period, and the subsequent autumn increase within a single inference task. Although posterior uncertainty increased and posterior predictive error was larger in this setting, the dominant temporal features of the epidemic trajectory were still reproduced, indicating that SBI remained informative even in a substantially higher-dimensional inference problem.

Due to the unidentifiability of the parameters that together make up the contact rate, the inferred reduction factors should be interpreted as changes in the effective contact rate rather than direct estimates of specific interventions. The reduction factors therefore summarize the combined effects of non-pharmaceutical interventions, behavioral adaptation, seasonal contact patterns, and other unobserved drivers of transmission~\cite{doi:10.1126/science.abb9789,Flaxman2020,ZUNKER2025116782}. As such, they are epidemiologically informative beyond trajectory reconstruction alone, since they can indicate shifts in transmission conditions and support timely model recalibration and assessment of epidemic dynamics. Consistent with this interpretation, the offset 0 window suggested sustained transmission suppression following the first epidemic wave. The offset 160 window indicated that transmission was already being reduced despite continued increases in ICU occupancy, reflecting delays between transmission changes and their impact on critical care demand. The offset 200 window was consistent with stronger transmission control and slower epidemic growth, although uncertainty increased for the final change point. Similar qualitative transmission patterns were recovered in the 201-day reconstruction.

Importantly, our evaluation was not limited to posterior predictive checks. Similar trajectory predictions can arise from different combinations of epidemiological and transmission parameters, particularly in partially non-identifiable compartmental models fitted to a single data set.
We therefore compared SBI and MCMC posteriors directly using marginal posterior distributions, Wasserstein distances, and KL divergences. This provides a stricter assessment of SBI than predictive performance alone. 
Across the 31-day windows, SBI recovered the major high-density regions of the MCMC reference posterior, although the SBI marginals were often smoother and broader. The broadening was especially visible in the later epidemic windows and the 201-day inference problem, consistent with the behavior  expected from amortized approximate inference, where the neural density estimator may trade posterior sharpness for smoother uncertainty representations, \cite{doi:10.1073/pnas.1912789117, lueckmann2021benchmarkingsimulationbasedinference}. Nevertheless, the dominant posterior structure and the posterior predictive behavior were preserved across the analyzed scenarios.

The broader SBI posteriors were also reflected in the posterior predictive checks, where SBI generally produced wider credible intervals than the MCMC reference despite similar posterior predictive means. One possible explanation is the difference in how uncertainty is incorporated during inference. The MCMC reference conditions parameter estimates through an explicit likelihood and observation error model, whereas the SBI approach was trained on deterministic model simulations and did not explicitly model observation noise. Consequently, SBI may retain posterior mass over a wider range of parameter combinations that generate plausible epidemic trajectories, leading to broader predictive uncertainty after propagation through the SECIR model. This effect was most apparent in the later epidemic windows, where uncertainty in transmission-related parameters is amplified by the nonlinear epidemic dynamics.

The comparison between time windows provides an additional consistency check of the inferred epidemiological parameters. Several parameters associated with disease progression showed substantial overlap across the different observation windows, despite representing distinct phases of the epidemic, suggesting that key disease-progression time scales were recovered similarly across changing epidemic conditions. Where differences between windows were observed, they appeared consistently in both SBI and MCMC, indicating that they likely reflected features of the data and model structure rather than artifacts of the SBI approximation.


At the same time, these differences should be interpreted with caution. Because the model is fitted to ICU occupancy alone and time-varying transmission was represented through effective contact-rate reduction factors, individual epidemiological parameters may be only partially identifiable and can be challenging to disentangle from changes in behavior, seasonality, reporting, healthcare practice, or residual model mismatch. Thus, changes in inferred symptomatic duration or symptomatic infectiousness should not be interpreted as direct evidence for biological changes in the disease progression. Rather, they should be understood as effective parameter changes within the calibrated model that help explain the observed ICU trajectory together with the inferred contact-rate parameters. 

The 201-day reconstruction provides further support for this interpretation. Because it covers the periods represented by the shorter 31-day windows within a single inference problem, it allows direct comparison between short- and long-window estimates. The long-window inference recovered transmission patterns that were qualitatively consistent with those identified in the shorter windows, although posterior uncertainty increased because of the larger number of inferred change points and stronger parameter correlations. Taken together, these findings indicate that the long-window inference provides a coherent, albeit less sharply identified, reconstruction of the epidemic phases captured by the shorter-window analyses.


SBI provided a substantial computational advantage over MCMC across both the short- and long-window inference problems. SBI reduced wall-clock inference time by approximately one order of magnitude for the 31-day windows and by more than two orders of magnitude for the 201-day reconstruction. The computational advantage became particularly pronounced in the higher-dimensional long-window setting, where MCMC required several hours and failed to achieve satisfactory convergence for all parameters. In contrast, SBI remained computationally tractable while maintaining broad agreement with the MCMC posterior and posterior predictive distributions. This reduction in runtime is particularly relevant for public-health applications in which epidemiological models must be recalibrated repeatedly as new data become available. Faster approximate posterior inference can facilitate more frequent model updates, rapid sensitivity analyses, and timely evaluation of alternative epidemic scenarios. 

However, our runtime comparison should be interpreted as a practical wall-clock comparison rather than a pure algorithmic benchmark, since the SBI workflow leveraged both CPU-based simulations and GPU-accelerated neural network training, whereas the MCMC reference was executed entirely on CPU hardware.

Our results also highlight several methodological considerations for applying SBI to epidemiological models. First, prior specification proved particularly important for the high-dimensional 201-day inference problem. Broad priors over multiple reduction parameters generated highly variable and partly implausible epidemic trajectories, reducing the relevance of the simulated training data and impairing posterior learning. This behavior is consistent with previous observations that SBI performance depends strongly on the quality of the prior predictive distribution and on sufficient coverage of posterior-relevant regions of parameter space during training \cite{papamakarios2018,greenberg2019}. 
Careful prior checking is therefore remains essential, particularly for long-window reconstruction problems involving many time-varying parameters

Second, we observed that increasing the number of simulations beyond 50{,}000 did not consistently improve posterior agreement or predictive accuracy in the 31-day inference problems. Instead, performance depended jointly on simulation budget, batch size, and inference complexity. Interestingly, the optimal configuration differed between the short- and long-window settings: while the 31-day analyses benefited from relatively large batch sizes, the substantially more complex 201-day inference required smaller batch sizes together with larger simulation budgets to achieve stable posterior recovery. One possible explanation is that smaller batches introduce additional stochasticity during optimization, which may improve generalization of the neural density estimator in higher-dimensional inference problems, consistent with observations from the deep-learning literature \cite{keskar2017on,masters2018revisitingsmallbatchtraining}. These findings suggest that hyperparameter selection for SBI in epidemiological applications may need to be adapted to the temporal scale and dimensionality of the inference problem rather than transferred directly between settings.

The study leaves several points for future research. First, the analysis was performed using a single mechanistic SECIR model and one national ICU occupancy time series. Further work is needed to assess the extent to which the conclusions generalize to other deterministic model structures, simulators, countries, pathogens, or surveillance settings.
Second, future studies could investigate whether combining ICU occupancy with additional epidemiological data sources improves parameter identifiability and posterior stability. 


Finally, the present analysis demonstrates retroperspective calibration rather than prospective forecasting. While the computational efficiency of SBI is promising for near-real-time epidemiological modelling, operational deployment would require evaluation in sequential forecasting settings with repeated updates as new observations become available.

In summary, SBI with NPE provided a fast and accurate approximation to MCMC-based Bayesian calibration for a mechanistic SECIR model fitted to German COVID-19 ICU occupancy data. Across multiple epidemic phases, SBI reproduced the observed ICU dynamics and recovered the dominant posterior structure of the MCMC reference while substantially reducing computational cost. Although posterior uncertainty increased in the more challenging long-window setting and performance remained sensitive to prior specification, SBI retained useful posterior information even in this higher-dimensional reconstruction problem. These findings support SBI as a promising approach for rapid calibration of mechanistic epidemiological models, provided that prior predictive checks, posterior validation, and careful consideration of parameter identifiability are incorporated into the inference workflow. Because SBI is an amortized inference method, its practical advantages are expected to be greatest in settings that require repeated inference under the same model and prior specification, where the initial simulation and training costs can be distributed across multiple analyses.



\section*{Acknowledgments}

This work was supported by the Initiative and Networking Fund of the Helmholtz Association (grant agreement number KA1-Co-08, Project LOKI-Pandemics). The authors gratefully acknowledge computing time on the supercomputer JURECA \cite{JURECA} at Forschungszentrum Jülich under grant no. loki.


\bibliography{references}

\clearpage
\section*{Supplementary information}
\addcontentsline{toc}{section}{Supplementary information}
\setcounter{section}{0}
\setcounter{figure}{0}
\setcounter{table}{0}
\renewcommand{\thesection}{S\arabic{section}}
\renewcommand{\thefigure}{S\arabic{figure}}
\renewcommand{\thetable}{S\arabic{table}}
\renewcommand{\theHsection}{S\arabic{section}}
\renewcommand{\theHfigure}{S\arabic{figure}}
\renewcommand{\theHtable}{S\arabic{table}}

\section{Mean Autoregressive Flows}\label{S1_Text}

Generally, normalizing flows learn the distribution of the data $X$ $p_X(x;\phi)$ through a mapping $f_{\phi}: \mathbb{R}^n\rightarrow \mathbb{R}^n$ between the latent variables $Z$ and the data $X$, where $\phi$ are the parameters of the mapping. Then, the change of variables formula is used
\begin{equation}
p_{X}(x;\phi)=p_{Z}(f^{-1}_{\phi}(x))\det|\frac{\partial f^{-1}_{\phi}}{\partial x}|,
\end{equation}
where $\det|\frac{\partial f^{-1}_{\phi}}{\partial x}|$ is a Jacobian of the transformation between $Z$ and $X$. Typically, $f_{\phi}$ is a chain of the transformations applied to a simple probability density function $p_Z$, such as the one of the normal distribution. The network is then train via minimizing the negative log-likelihood
\begin{equation}
\mathcal{L}=-\log{p_X(x)}=-\log{p_{Z}(f^{-1}_{\phi}(x))}-\log{\det|\frac{\partial f^{-1}_{\phi}}{\partial x}|}
\end{equation}

MAFs model the probability density function of the data in an autoregressive way $p_X(x)=\prod_{i=1}^np(x_{i}|x_{1:i-1})$, where $x=(x_1,\dots,x_n)$ and $x_{1:i}=(x_1,\dots,x_i)$. 
\begin{equation}
p(x_i|x_{1:i})\sim \mathcal{N}(x_i|m_{i,\phi},\exp{s_{i,\phi}}), m_{i,\phi}=m_{i,\phi}(x_{1:i-1}), s_{i,\phi}=s_{i,\phi}(x_{1:i-1})
\end{equation}

where $m_{i,\phi}$ and $s_{i,\phi}$ are neural networks with shared weights $\phi$ which is performed by Masked Autoencoder for Density Estimation (MADE) \cite{germain2015mademaskedautoencoderdistribution}. The latter applies binary masks on weight matrices in such a way that each output $i$ depends only on the inputs $1:i$. Therefore the transformation from $Z$ to $X$ is as follows
\begin{equation}
x_i=z_i\exp{s_{i,\phi}}+m_{i,\phi},\quad z_i\sim \mathcal{N}(0,1)
\end{equation}

In the neural posterior estimation setting considered in this work, the
variables transformed by the flow correspond to the epidemiological parameters
$\theta$, while the simulated observations $\hat X$ are provided as
conditioning information. These conditioning variables are supplied to the
MADE network and influence the functions $m_{i,\phi}$ and $s_{i,\phi}$ that
parameterize the autoregressive transformations, while not themselves being
transformed by the flow. Consequently, the learned distribution is

\begin{equation}
p(\theta|\hat X)
=
\prod_{i=1}^{n}
p(\theta_i|\theta_{1:i-1},\hat X).
\end{equation}

\section{SNPE-C Training Objective}\label{S2_Text}
In SBI framework the training loss consists of two components: an atomic loss, which performs contrastive training on proposal-sampled parameter atoms, and an additional maximum-likelihood (MLE) term. The MLE term here acts as a regularizer by encouraging the density estimator to assign high probability to prior-drawn samples, thereby reducing density leakage outside the bounded support of the prior. Note, that the function to learn here is $p_X(\theta;\phi|x)$, corresponding to the posterior distribution of the parameters $\theta$ given the simulated data $x$. Atomic loss reads as follows
\begin{equation}
\mathcal{L}_{atomic}=-\log\frac{p(x;\phi|\theta^k)}{\sum_{j=1}^{K}p_X(x;\phi|\theta^j)},
\end{equation}
where $\{\theta^1,\dots,\theta^K\}$ are all sets of parameters $\theta$ generated during training, while $\theta^k$ is the one underlying the simulated trajectory $x$.
\begin{equation}
\mathcal{L}_{MLE}=-\log{p_{X}(\theta;\phi|x)}=-\log{p_{Z}(f^{-1}_{\phi}(\theta;x))}-\log{\det|\frac{\partial f^{-1}_{\phi}}{\partial \theta}|}
\end{equation}

The final loss being
\begin{equation}
\mathcal{L}=\mathcal{L}_{atomic}+\mathcal{L}_{MLE}
\end{equation}
\newpage
\begin{figure}[H]
    \centering
    \makebox[\textwidth][c]{ 
        \includegraphics[width=0.49\linewidth]{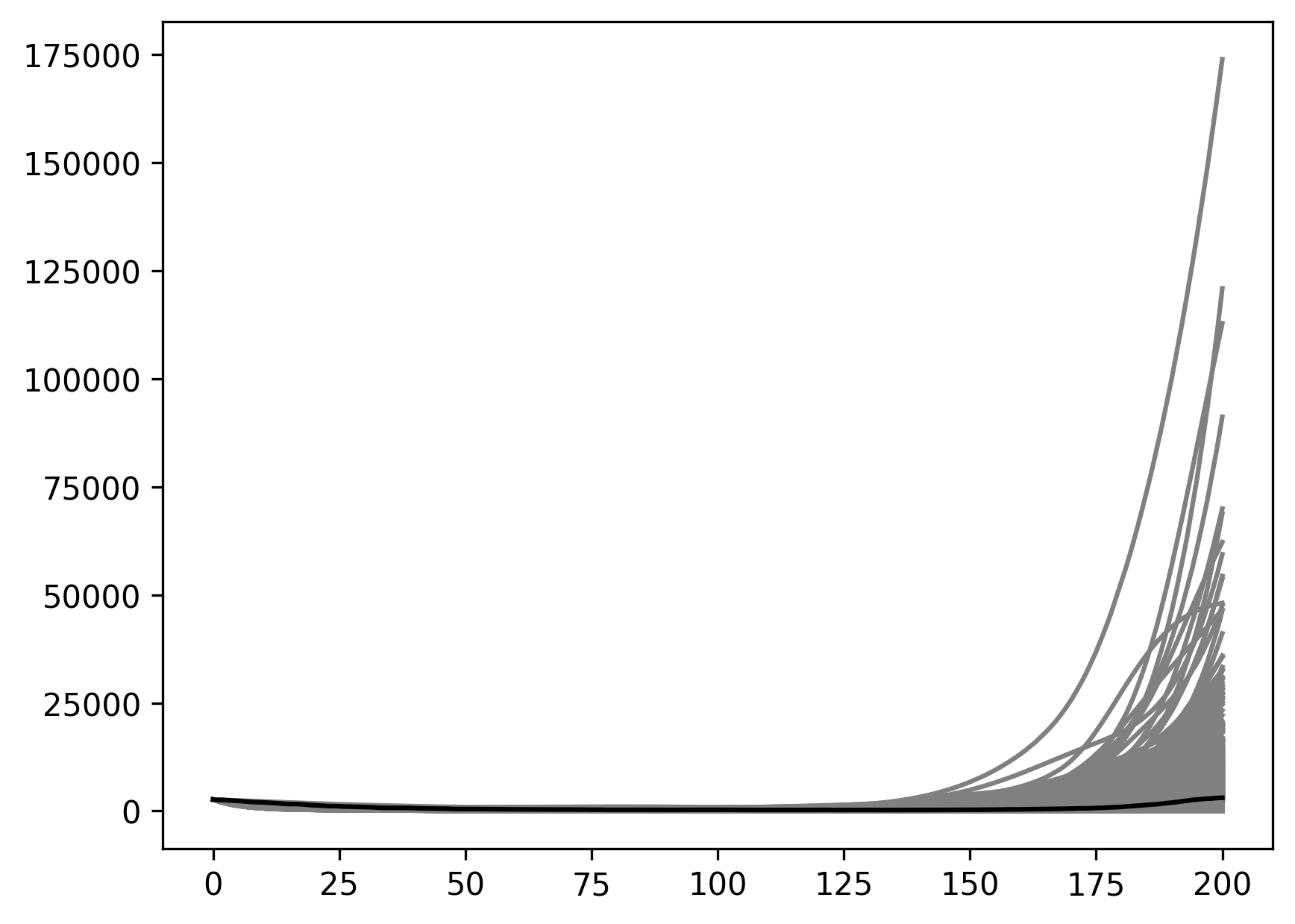}
        \includegraphics[width=0.49\linewidth]{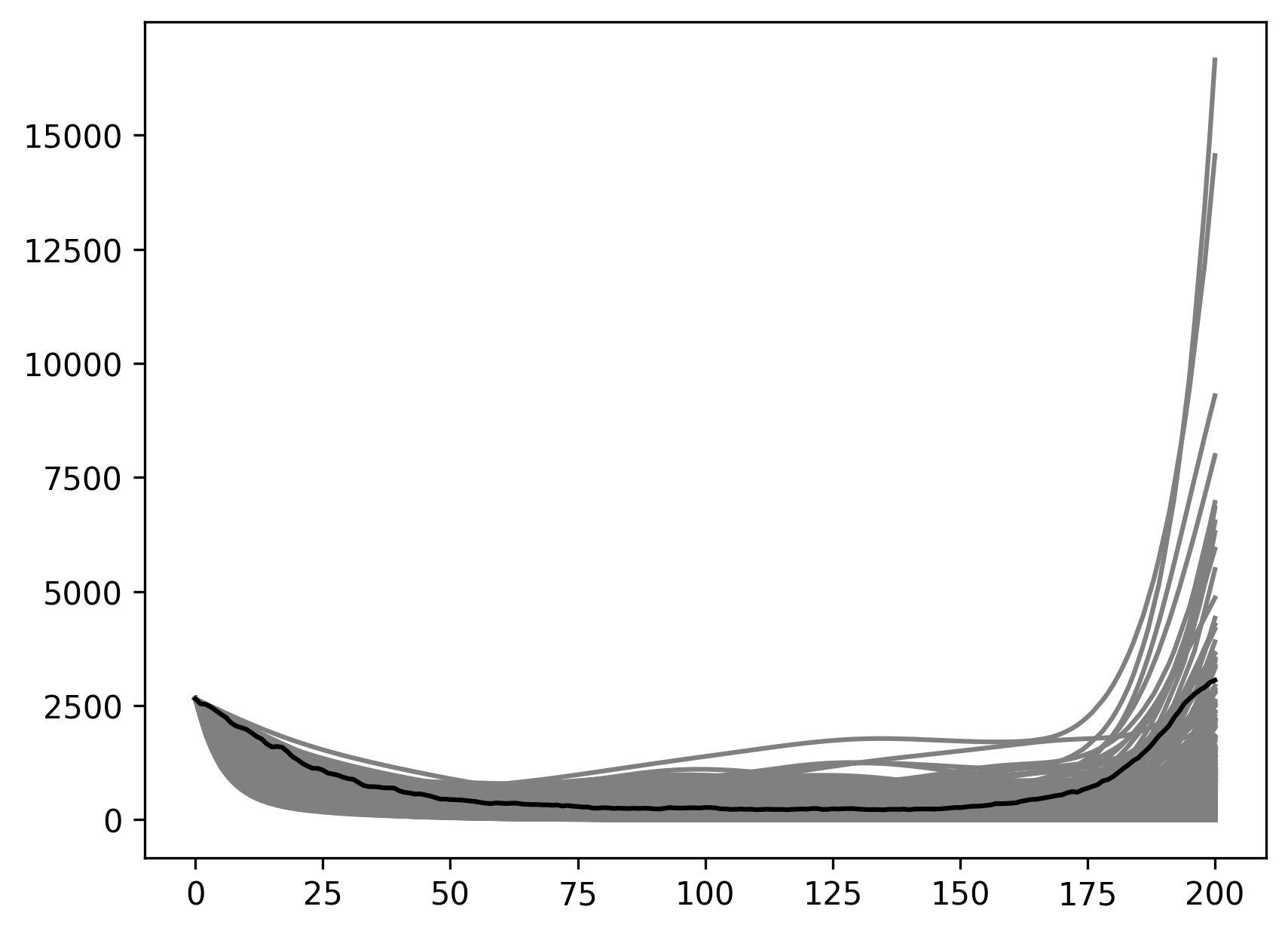}
    }
    \caption{\textbf{Prior predictive simulations for the 200-day inference setting under two alternative change-point prior specifications.} 
    Each panel shows 10{,}000 ICU occupancy trajectories (vertical axis) over 200 days (horizontal axis) generated from parameters sampled from the prior distribution and propagated through the forward simulator. The observed trajectory (black curve) is overlaid for reference. Broader prior bounds (left, Spec A in Table~\ref{S1_Table}) result in highly dispersed and partially implausible epidemic dynamics, whereas tighter bounds (right, Spec B in Table~\ref{S1_Table}) concentrate prior mass around epidemiologically plausible trajectories while retaining sufficient variability}
    \label{S1_Fig}
\end{figure}

\newpage
\begin{figure}[H]
    \centering
    
        \includegraphics[width=0.67\linewidth]{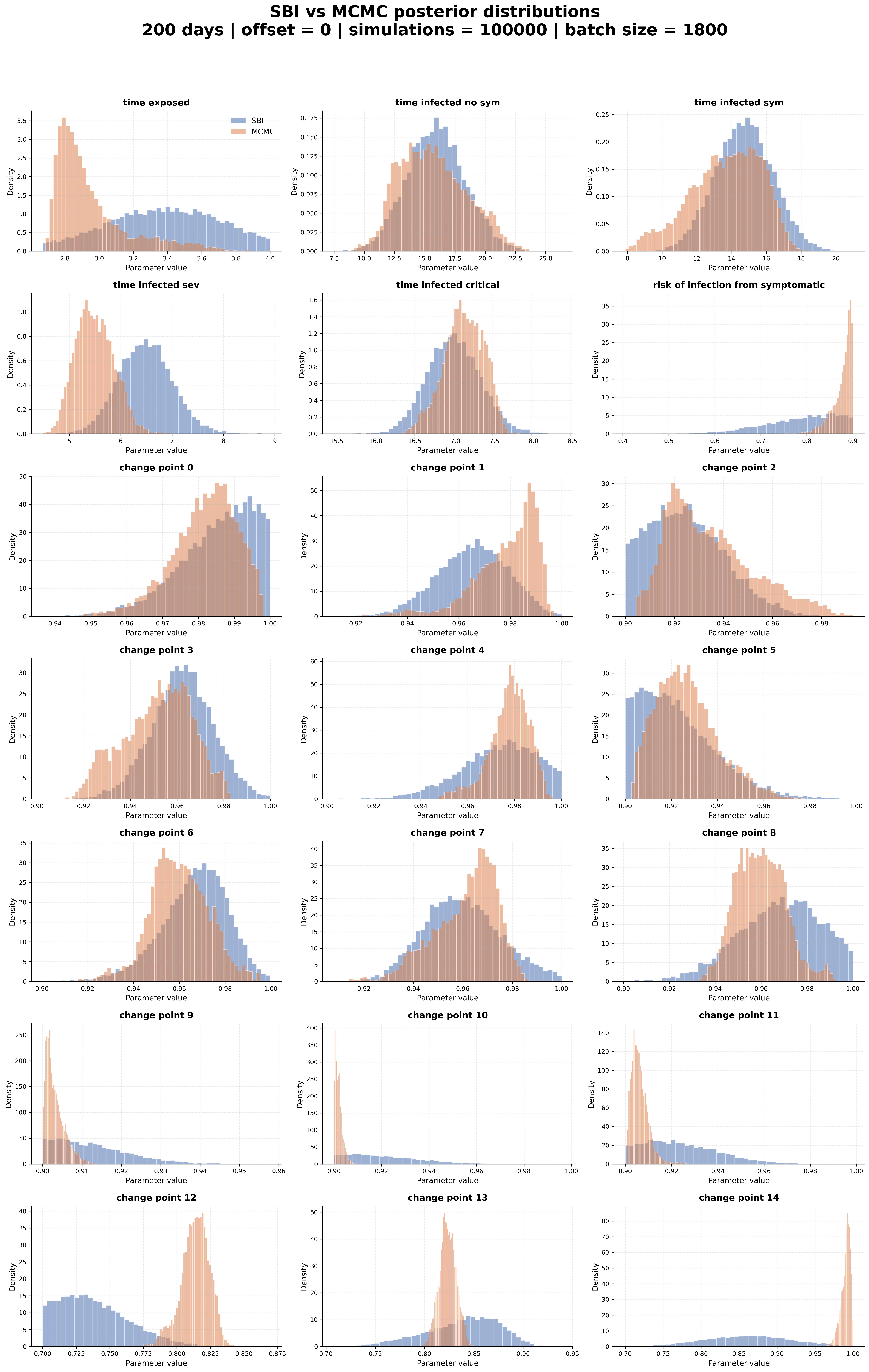}

    \caption{{\bf Comparison of marginal posterior distributions inferred using simulation-based inference (SBI, blue) and Markov chain Monte Carlo (MCMC, orange) for the extended 200-day inference window starting on April 24, 2020.} The panels show posterior distributions for epidemiological parameters and all inferred change-point parameters. Despite the substantially increased dimensionality of the inference problem, SBI recovered posterior modes and overall parameter trends broadly consistent with the corresponding MCMC posteriors, while generally producing smoother and broader marginal distributions.}
    \label{S2_Fig}
\end{figure}

\begin{table}[H]
\centering
\caption{\textbf{Prior bounds for contact-reduction (damping) parameters in the 201-day inference problem.}
Lower and upper bounds of the uniform prior distributions assigned to the 15 time-varying contact-reduction parameters ($r_0,\dots,r_{14}$). Specification A corresponds to the initial prior configuration, while Specification B denotes the revised prior used after prior predictive checking to improve the plausibility of simulated epidemic trajectories and reduce prior support for unrealistic transmission dynamics, as show in Fig.~\ref{S1_Fig}.}
\label{S1_Table}
\resizebox{\textwidth}{!}{%
\begin{tabular}{c c c c c}
\hline
\textbf{reduction factor} & \textbf{Lower (Spec A)} & \textbf{Upper (Spec A)} & \textbf{Lower (Spec B)} & \textbf{Upper (Spec B)} \\
\hline
0  & 0.9 & 1.0 & 0.9 & 1.0 \\
1  & 0.9 & 1.0 & 0.9 & 1.0 \\
2  & 0.9 & 1.0 & 0.9 & 1.0 \\
3  & 0.9 & 1.0 & 0.9 & 1.0 \\
4  & 0.9 & 1.0 & 0.9 & 1.0 \\
5  & 0.9 & 1.0 & 0.9 & 1.0 \\
6  & 0.9 & 1.0 & 0.9 & 1.0 \\
7  & 0.9 & 1.0 & 0.9 & 1.0 \\
8  & 0.8 & 1.0 & 0.9 & 1.0 \\
9  & 0.8 & 1.0 & 0.9 & 1.0 \\
10 & 0.8 & 1.0 & 0.9 & 1.0 \\
11 & 0.8 & 1.0 & 0.9 & 1.0 \\
12 & 0.7 & 1.0 & 0.7 & 1.0 \\
13 & 0.7 & 1.0 & 0.7 & 1.0 \\
14 & 0.7 & 1.0 & 0.7 & 1.0 \\
\hline
\end{tabular}%
}
\end{table}

\begin{table}[htbp]
\centering
\small
\caption{
\textbf{MCMC runtime details and diagnostics for all runs.} For each inference setting, we report the number of chains and samples, acceptance rate, runtime, and maximum $\hat{R}$ value. The reported maximum $\hat{R}$ and threshold counts refer to inferred model parameters. 
The 30 day runs converged nicely, with just the 31 day, offset 160 run having a single parameter with $\hat{R}=1.012>1.01$, which can effectively be interpreted as convergence. Conversely, for the 200 day run no parameters converged despite the long runtime and large sample budget. As shown in Table~\ref{S3_Table}, the $\hat{R}$ values for the posterior predictive values are lower, between $1.009$ and $1.250$, indicating that the $16$ chains mostly agree on the posterior predictive even though they do not agree on the parameters. To keep computations light, we thinned the 201 days MCMC run by a factor of 100, significantly reducing memory footprint.
}
\label{S2_Table}
\small
\setlength{\tabcolsep}{3pt}
\resizebox{\textwidth}{!}{%
\begin{tabular}{@{}lccccccc@{}}
\toprule
Run & Chains & Draws / chain & Total samples & Runtime & Acceptance rate & Max parameter $\hat{R}$ & Parameters with $\hat{R}>1.01$ \\
\midrule
31 days, offset 0 & 16 & 100\,000 & 1\,600\,000 & $1.08\text{e}3$s & 0.391 & 1.001 & 0 \\
31 days, offset 160 & 16 & 100\,000 & 1\,600\,000 & $1.09\text{e}3$s & 0.394 & 1.012 & 1 \\
31 days, offset 200 & 16 & 100\,000 & 1\,600\,000 & $1.07\text{e}3$s & 0.418 & 1.008 & 0 \\
201 days, offset 0 & 16 & 1\,000\,000 & 16\,000\,000 & $1.91\text{e}4$s & 0.526 & 1.606 & 21 \\
\bottomrule
\end{tabular}%
}
\end{table}

\begingroup
\scriptsize
\setlength{\tabcolsep}{3pt}
\begin{longtable}{lrrrr}
\caption{
\textbf{Rank-normalized $\hat{R}$ values for all parameters and posterior predictive ICU occupancy.} The predictive quantities are reported as derived diagnostics of chain agreement for the model outputs. For the first day of simulation (day 0, day 160, day 200), each of the runs used the initialization process described in materials and methods.
The 31 day inference windows show good convergence. For the 200 day run, mixing was harder for the parameters than for the posterior predictive values and despite the long runtime, MCMC did not achieve convergence.}\label{S3_Table}\\
\toprule
Quantity & 31 days, offset 0 & 31 days, offset 160 & 31 days, offset 200 & 201 days \\
\midrule
\endfirsthead
\toprule
Quantity & 31 days, offset 0 & 31 days, offset 160 & 31 days, offset 200 & 201 days \\
\midrule
\endhead
\midrule
\multicolumn{5}{r}{Continued on next page} \\
\endfoot
\bottomrule
\endlastfoot
\addlinespace
\multicolumn{5}{l}{\textit{Inferred model parameters}} \\
reduction factor 0 & 1.001 & 1.010 & 1.006 & \textbf{1.554} \\
reduction factor 1 & 1.000 & 1.003 & 1.004 & \textbf{1.284} \\
reduction factor 2 & -- & -- & -- & \textbf{1.449} \\
reduction factor 3 & -- & -- & -- & \textbf{1.217} \\
reduction factor 4 & -- & -- & -- & \textbf{1.380} \\
reduction factor 5 & -- & -- & -- & \textbf{1.543} \\
reduction factor 6 & -- & -- & -- & \textbf{1.606} \\
reduction factor 7 & -- & -- & -- & \textbf{1.544} \\
reduction factor 8 & -- & -- & -- & \textbf{1.309} \\
reduction factor 9 & -- & -- & -- & \textbf{1.160} \\
reduction factor 10 & -- & -- & -- & \textbf{1.066} \\
reduction factor 11 & -- & -- & -- & \textbf{1.126} \\
reduction factor 12 & -- & -- & -- & \textbf{1.214} \\
reduction factor 13 & -- & -- & -- & \textbf{1.080} \\
reduction factor 14 & -- & -- & -- & \textbf{1.155} \\
risk of infection from symptomatic & 1.000 & \textbf{1.012} & 1.008 & \textbf{1.317} \\
time exposed & 1.000 & 1.002 & 1.002 & \textbf{1.228} \\
time infected critical & 1.001 & 1.002 & 1.004 & \textbf{1.369} \\
time infected no sym & 1.001 & 1.003 & 1.003 & \textbf{1.401} \\
time infected sev & 1.000 & 1.004 & 1.004 & \textbf{1.067} \\
time infected sym & 1.000 & 1.003 & 1.003 & \textbf{1.470} \\
\addlinespace
\multicolumn{5}{l}{\textit{Posterior predictive ICU occupancy}} \\
predicted ICU occupancy day 0 & -- & -- & -- & -- \\
predicted ICU occupancy day 1 & 2.000 & -- & -- & \textbf{1.106} \\
predicted ICU occupancy day 2 & 1.000 & -- & -- & \textbf{1.091} \\
predicted ICU occupancy day 3 & 1.000 & -- & -- & \textbf{1.069} \\
predicted ICU occupancy day 4 & 1.000 & -- & -- & \textbf{1.049} \\
predicted ICU occupancy day 5 & 1.000 & -- & -- & \textbf{1.033} \\
predicted ICU occupancy day 6 & 1.001 & -- & -- & \textbf{1.022} \\
predicted ICU occupancy day 7 & 1.001 & -- & -- & \textbf{1.017} \\
predicted ICU occupancy day 8 & 1.001 & -- & -- & \textbf{1.021} \\
predicted ICU occupancy day 9 & 1.001 & -- & -- & \textbf{1.031} \\
predicted ICU occupancy day 10 & 1.001 & -- & -- & \textbf{1.046} \\
predicted ICU occupancy day 11 & 1.001 & -- & -- & \textbf{1.062} \\
predicted ICU occupancy day 12 & 1.001 & -- & -- & \textbf{1.077} \\
predicted ICU occupancy day 13 & 1.001 & -- & -- & \textbf{1.088} \\
predicted ICU occupancy day 14 & 1.001 & -- & -- & \textbf{1.094} \\
predicted ICU occupancy day 15 & 1.001 & -- & -- & \textbf{1.095} \\
predicted ICU occupancy day 16 & 1.001 & -- & -- & \textbf{1.093} \\
predicted ICU occupancy day 17 & 1.001 & -- & -- & \textbf{1.088} \\
predicted ICU occupancy day 18 & 1.000 & -- & -- & \textbf{1.081} \\
predicted ICU occupancy day 19 & 1.000 & -- & -- & \textbf{1.073} \\
predicted ICU occupancy day 20 & 1.000 & -- & -- & \textbf{1.064} \\
predicted ICU occupancy day 21 & 1.000 & -- & -- & \textbf{1.056} \\
predicted ICU occupancy day 22 & 1.000 & -- & -- & \textbf{1.050} \\
predicted ICU occupancy day 23 & 1.000 & -- & -- & \textbf{1.044} \\
predicted ICU occupancy day 24 & 1.000 & -- & -- & \textbf{1.041} \\
predicted ICU occupancy day 25 & 1.000 & -- & -- & \textbf{1.041} \\
predicted ICU occupancy day 26 & 1.000 & -- & -- & \textbf{1.041} \\
predicted ICU occupancy day 27 & 1.000 & -- & -- & \textbf{1.043} \\
predicted ICU occupancy day 28 & 1.000 & -- & -- & \textbf{1.044} \\
predicted ICU occupancy day 29 & 1.000 & -- & -- & \textbf{1.047} \\
predicted ICU occupancy day 30 & 1.000 & -- & -- & \textbf{1.051} \\
predicted ICU occupancy day 31 & -- & -- & -- & \textbf{1.055} \\
predicted ICU occupancy day 32 & -- & -- & -- & \textbf{1.059} \\
predicted ICU occupancy day 33 & -- & -- & -- & \textbf{1.062} \\
predicted ICU occupancy day 34 & -- & -- & -- & \textbf{1.068} \\
predicted ICU occupancy day 35 & -- & -- & -- & \textbf{1.073} \\
predicted ICU occupancy day 36 & -- & -- & -- & \textbf{1.079} \\
predicted ICU occupancy day 37 & -- & -- & -- & \textbf{1.086} \\
predicted ICU occupancy day 38 & -- & -- & -- & \textbf{1.092} \\
predicted ICU occupancy day 39 & -- & -- & -- & \textbf{1.099} \\
predicted ICU occupancy day 40 & -- & -- & -- & \textbf{1.104} \\
predicted ICU occupancy day 41 & -- & -- & -- & \textbf{1.109} \\
predicted ICU occupancy day 42 & -- & -- & -- & \textbf{1.111} \\
predicted ICU occupancy day 43 & -- & -- & -- & \textbf{1.111} \\
predicted ICU occupancy day 44 & -- & -- & -- & \textbf{1.109} \\
predicted ICU occupancy day 45 & -- & -- & -- & \textbf{1.103} \\
predicted ICU occupancy day 46 & -- & -- & -- & \textbf{1.094} \\
predicted ICU occupancy day 47 & -- & -- & -- & \textbf{1.083} \\
predicted ICU occupancy day 48 & -- & -- & -- & \textbf{1.070} \\
predicted ICU occupancy day 49 & -- & -- & -- & \textbf{1.058} \\
predicted ICU occupancy day 50 & -- & -- & -- & \textbf{1.047} \\
predicted ICU occupancy day 51 & -- & -- & -- & \textbf{1.040} \\
predicted ICU occupancy day 52 & -- & -- & -- & \textbf{1.037} \\
predicted ICU occupancy day 53 & -- & -- & -- & \textbf{1.035} \\
predicted ICU occupancy day 54 & -- & -- & -- & \textbf{1.038} \\
predicted ICU occupancy day 55 & -- & -- & -- & \textbf{1.040} \\
predicted ICU occupancy day 56 & -- & -- & -- & \textbf{1.044} \\
predicted ICU occupancy day 57 & -- & -- & -- & \textbf{1.048} \\
predicted ICU occupancy day 58 & -- & -- & -- & \textbf{1.052} \\
predicted ICU occupancy day 59 & -- & -- & -- & \textbf{1.055} \\
predicted ICU occupancy day 60 & -- & -- & -- & \textbf{1.059} \\
predicted ICU occupancy day 61 & -- & -- & -- & \textbf{1.061} \\
predicted ICU occupancy day 62 & -- & -- & -- & \textbf{1.063} \\
predicted ICU occupancy day 63 & -- & -- & -- & \textbf{1.065} \\
predicted ICU occupancy day 64 & -- & -- & -- & \textbf{1.065} \\
predicted ICU occupancy day 65 & -- & -- & -- & \textbf{1.065} \\
predicted ICU occupancy day 66 & -- & -- & -- & \textbf{1.064} \\
predicted ICU occupancy day 67 & -- & -- & -- & \textbf{1.062} \\
predicted ICU occupancy day 68 & -- & -- & -- & \textbf{1.061} \\
predicted ICU occupancy day 69 & -- & -- & -- & \textbf{1.060} \\
predicted ICU occupancy day 70 & -- & -- & -- & \textbf{1.061} \\
predicted ICU occupancy day 71 & -- & -- & -- & \textbf{1.064} \\
predicted ICU occupancy day 72 & -- & -- & -- & \textbf{1.070} \\
predicted ICU occupancy day 73 & -- & -- & -- & \textbf{1.078} \\
predicted ICU occupancy day 74 & -- & -- & -- & \textbf{1.088} \\
predicted ICU occupancy day 75 & -- & -- & -- & \textbf{1.098} \\
predicted ICU occupancy day 76 & -- & -- & -- & \textbf{1.107} \\
predicted ICU occupancy day 77 & -- & -- & -- & \textbf{1.114} \\
predicted ICU occupancy day 78 & -- & -- & -- & \textbf{1.117} \\
predicted ICU occupancy day 79 & -- & -- & -- & \textbf{1.114} \\
predicted ICU occupancy day 80 & -- & -- & -- & \textbf{1.108} \\
predicted ICU occupancy day 81 & -- & -- & -- & \textbf{1.097} \\
predicted ICU occupancy day 82 & -- & -- & -- & \textbf{1.083} \\
predicted ICU occupancy day 83 & -- & -- & -- & \textbf{1.069} \\
predicted ICU occupancy day 84 & -- & -- & -- & \textbf{1.057} \\
predicted ICU occupancy day 85 & -- & -- & -- & \textbf{1.051} \\
predicted ICU occupancy day 86 & -- & -- & -- & \textbf{1.054} \\
predicted ICU occupancy day 87 & -- & -- & -- & \textbf{1.069} \\
predicted ICU occupancy day 88 & -- & -- & -- & \textbf{1.095} \\
predicted ICU occupancy day 89 & -- & -- & -- & \textbf{1.128} \\
predicted ICU occupancy day 90 & -- & -- & -- & \textbf{1.163} \\
predicted ICU occupancy day 91 & -- & -- & -- & \textbf{1.195} \\
predicted ICU occupancy day 92 & -- & -- & -- & \textbf{1.220} \\
predicted ICU occupancy day 93 & -- & -- & -- & \textbf{1.238} \\
predicted ICU occupancy day 94 & -- & -- & -- & \textbf{1.247} \\
predicted ICU occupancy day 95 & -- & -- & -- & \textbf{1.250} \\
predicted ICU occupancy day 96 & -- & -- & -- & \textbf{1.247} \\
predicted ICU occupancy day 97 & -- & -- & -- & \textbf{1.241} \\
predicted ICU occupancy day 98 & -- & -- & -- & \textbf{1.231} \\
predicted ICU occupancy day 99 & -- & -- & -- & \textbf{1.219} \\
predicted ICU occupancy day 100 & -- & -- & -- & \textbf{1.207} \\
predicted ICU occupancy day 101 & -- & -- & -- & \textbf{1.194} \\
predicted ICU occupancy day 102 & -- & -- & -- & \textbf{1.182} \\
predicted ICU occupancy day 103 & -- & -- & -- & \textbf{1.171} \\
predicted ICU occupancy day 104 & -- & -- & -- & \textbf{1.161} \\
predicted ICU occupancy day 105 & -- & -- & -- & \textbf{1.151} \\
predicted ICU occupancy day 106 & -- & -- & -- & \textbf{1.142} \\
predicted ICU occupancy day 107 & -- & -- & -- & \textbf{1.133} \\
predicted ICU occupancy day 108 & -- & -- & -- & \textbf{1.123} \\
predicted ICU occupancy day 109 & -- & -- & -- & \textbf{1.112} \\
predicted ICU occupancy day 110 & -- & -- & -- & \textbf{1.099} \\
predicted ICU occupancy day 111 & -- & -- & -- & \textbf{1.084} \\
predicted ICU occupancy day 112 & -- & -- & -- & \textbf{1.069} \\
predicted ICU occupancy day 113 & -- & -- & -- & \textbf{1.053} \\
predicted ICU occupancy day 114 & -- & -- & -- & \textbf{1.040} \\
predicted ICU occupancy day 115 & -- & -- & -- & \textbf{1.030} \\
predicted ICU occupancy day 116 & -- & -- & -- & \textbf{1.025} \\
predicted ICU occupancy day 117 & -- & -- & -- & \textbf{1.024} \\
predicted ICU occupancy day 118 & -- & -- & -- & \textbf{1.026} \\
predicted ICU occupancy day 119 & -- & -- & -- & \textbf{1.032} \\
predicted ICU occupancy day 120 & -- & -- & -- & \textbf{1.039} \\
predicted ICU occupancy day 121 & -- & -- & -- & \textbf{1.046} \\
predicted ICU occupancy day 122 & -- & -- & -- & \textbf{1.053} \\
predicted ICU occupancy day 123 & -- & -- & -- & \textbf{1.060} \\
predicted ICU occupancy day 124 & -- & -- & -- & \textbf{1.065} \\
predicted ICU occupancy day 125 & -- & -- & -- & \textbf{1.069} \\
predicted ICU occupancy day 126 & -- & -- & -- & \textbf{1.072} \\
predicted ICU occupancy day 127 & -- & -- & -- & \textbf{1.074} \\
predicted ICU occupancy day 128 & -- & -- & -- & \textbf{1.076} \\
predicted ICU occupancy day 129 & -- & -- & -- & \textbf{1.077} \\
predicted ICU occupancy day 130 & -- & -- & -- & \textbf{1.077} \\
predicted ICU occupancy day 131 & -- & -- & -- & \textbf{1.078} \\
predicted ICU occupancy day 132 & -- & -- & -- & \textbf{1.079} \\
predicted ICU occupancy day 133 & -- & -- & -- & \textbf{1.082} \\
predicted ICU occupancy day 134 & -- & -- & -- & \textbf{1.084} \\
predicted ICU occupancy day 135 & -- & -- & -- & \textbf{1.087} \\
predicted ICU occupancy day 136 & -- & -- & -- & \textbf{1.090} \\
predicted ICU occupancy day 137 & -- & -- & -- & \textbf{1.095} \\
predicted ICU occupancy day 138 & -- & -- & -- & \textbf{1.101} \\
predicted ICU occupancy day 139 & -- & -- & -- & \textbf{1.104} \\
predicted ICU occupancy day 140 & -- & -- & -- & \textbf{1.108} \\
predicted ICU occupancy day 141 & -- & -- & -- & \textbf{1.112} \\
predicted ICU occupancy day 142 & -- & -- & -- & \textbf{1.115} \\
predicted ICU occupancy day 143 & -- & -- & -- & \textbf{1.117} \\
predicted ICU occupancy day 144 & -- & -- & -- & \textbf{1.117} \\
predicted ICU occupancy day 145 & -- & -- & -- & \textbf{1.115} \\
predicted ICU occupancy day 146 & -- & -- & -- & \textbf{1.115} \\
predicted ICU occupancy day 147 & -- & -- & -- & \textbf{1.113} \\
predicted ICU occupancy day 148 & -- & -- & -- & \textbf{1.109} \\
predicted ICU occupancy day 149 & -- & -- & -- & \textbf{1.104} \\
predicted ICU occupancy day 150 & -- & -- & -- & \textbf{1.097} \\
predicted ICU occupancy day 151 & -- & -- & -- & \textbf{1.089} \\
predicted ICU occupancy day 152 & -- & -- & -- & \textbf{1.080} \\
predicted ICU occupancy day 153 & -- & -- & -- & \textbf{1.071} \\
predicted ICU occupancy day 154 & -- & -- & -- & \textbf{1.062} \\
predicted ICU occupancy day 155 & -- & -- & -- & \textbf{1.052} \\
predicted ICU occupancy day 156 & -- & -- & -- & \textbf{1.042} \\
predicted ICU occupancy day 157 & -- & -- & -- & \textbf{1.033} \\
predicted ICU occupancy day 158 & -- & -- & -- & \textbf{1.026} \\
predicted ICU occupancy day 159 & -- & -- & -- & \textbf{1.020} \\
predicted ICU occupancy day 160 & -- & -- & -- & \textbf{1.018} \\
predicted ICU occupancy day 161 & -- & 1.004 & -- & \textbf{1.019} \\
predicted ICU occupancy day 162 & -- & 1.004 & -- & \textbf{1.023} \\
predicted ICU occupancy day 163 & -- & 1.004 & -- & \textbf{1.029} \\
predicted ICU occupancy day 164 & -- & 1.003 & -- & \textbf{1.037} \\
predicted ICU occupancy day 165 & -- & 1.003 & -- & \textbf{1.044} \\
predicted ICU occupancy day 166 & -- & 1.002 & -- & \textbf{1.057} \\
predicted ICU occupancy day 167 & -- & 1.002 & -- & \textbf{1.064} \\
predicted ICU occupancy day 168 & -- & 1.002 & -- & \textbf{1.070} \\
predicted ICU occupancy day 169 & -- & 1.002 & -- & \textbf{1.076} \\
predicted ICU occupancy day 170 & -- & 1.002 & -- & \textbf{1.080} \\
predicted ICU occupancy day 171 & -- & 1.002 & -- & \textbf{1.080} \\
predicted ICU occupancy day 172 & -- & 1.002 & -- & \textbf{1.079} \\
predicted ICU occupancy day 173 & -- & 1.002 & -- & \textbf{1.077} \\
predicted ICU occupancy day 174 & -- & 1.002 & -- & \textbf{1.072} \\
predicted ICU occupancy day 175 & -- & 1.003 & -- & \textbf{1.064} \\
predicted ICU occupancy day 176 & -- & 1.003 & -- & \textbf{1.054} \\
predicted ICU occupancy day 177 & -- & 1.002 & -- & \textbf{1.043} \\
predicted ICU occupancy day 178 & -- & 1.002 & -- & \textbf{1.032} \\
predicted ICU occupancy day 179 & -- & 1.002 & -- & \textbf{1.025} \\
predicted ICU occupancy day 180 & -- & 1.002 & -- & \textbf{1.023} \\
predicted ICU occupancy day 181 & -- & 1.001 & -- & \textbf{1.020} \\
predicted ICU occupancy day 182 & -- & 1.001 & -- & \textbf{1.018} \\
predicted ICU occupancy day 183 & -- & 1.001 & -- & \textbf{1.016} \\
predicted ICU occupancy day 184 & -- & 1.001 & -- & \textbf{1.015} \\
predicted ICU occupancy day 185 & -- & 1.001 & -- & \textbf{1.016} \\
predicted ICU occupancy day 186 & -- & 1.001 & -- & \textbf{1.015} \\
predicted ICU occupancy day 187 & -- & 1.000 & -- & \textbf{1.016} \\
predicted ICU occupancy day 188 & -- & 1.000 & -- & \textbf{1.019} \\
predicted ICU occupancy day 189 & -- & 1.001 & -- & \textbf{1.021} \\
predicted ICU occupancy day 190 & -- & 1.001 & -- & \textbf{1.023} \\
predicted ICU occupancy day 191 & -- & -- & -- & \textbf{1.022} \\
predicted ICU occupancy day 192 & -- & -- & -- & \textbf{1.019} \\
predicted ICU occupancy day 193 & -- & -- & -- & \textbf{1.014} \\
predicted ICU occupancy day 194 & -- & -- & -- & 1.009 \\
predicted ICU occupancy day 195 & -- & -- & -- & 1.006 \\
predicted ICU occupancy day 196 & -- & -- & -- & 1.005 \\
predicted ICU occupancy day 197 & -- & -- & -- & 1.009 \\
predicted ICU occupancy day 198 & -- & -- & -- & \textbf{1.016} \\
predicted ICU occupancy day 199 & -- & -- & -- & \textbf{1.025} \\
predicted ICU occupancy day 200 & -- & -- & -- & \textbf{1.034} \\
predicted ICU occupancy day 201 & -- & -- & 1.003 & -- \\
predicted ICU occupancy day 202 & -- & -- & 1.003 & -- \\
predicted ICU occupancy day 203 & -- & -- & 1.003 & -- \\
predicted ICU occupancy day 204 & -- & -- & 1.003 & -- \\
predicted ICU occupancy day 205 & -- & -- & 1.003 & -- \\
predicted ICU occupancy day 206 & -- & -- & 1.003 & -- \\
predicted ICU occupancy day 207 & -- & -- & 1.003 & -- \\
predicted ICU occupancy day 208 & -- & -- & 1.002 & -- \\
predicted ICU occupancy day 209 & -- & -- & 1.002 & -- \\
predicted ICU occupancy day 210 & -- & -- & 1.002 & -- \\
predicted ICU occupancy day 211 & -- & -- & 1.002 & -- \\
predicted ICU occupancy day 212 & -- & -- & 1.002 & -- \\
predicted ICU occupancy day 213 & -- & -- & 1.002 & -- \\
predicted ICU occupancy day 214 & -- & -- & 1.003 & -- \\
predicted ICU occupancy day 215 & -- & -- & 1.003 & -- \\
predicted ICU occupancy day 216 & -- & -- & 1.003 & -- \\
predicted ICU occupancy day 217 & -- & -- & 1.003 & -- \\
predicted ICU occupancy day 218 & -- & -- & 1.002 & -- \\
predicted ICU occupancy day 219 & -- & -- & 1.002 & -- \\
predicted ICU occupancy day 220 & -- & -- & 1.001 & -- \\
predicted ICU occupancy day 221 & -- & -- & 1.001 & -- \\
predicted ICU occupancy day 222 & -- & -- & 1.001 & -- \\
predicted ICU occupancy day 223 & -- & -- & 1.001 & -- \\
predicted ICU occupancy day 224 & -- & -- & 1.001 & -- \\
predicted ICU occupancy day 225 & -- & -- & 1.001 & -- \\
predicted ICU occupancy day 226 & -- & -- & 1.001 & -- \\
predicted ICU occupancy day 227 & -- & -- & 1.001 & -- \\
predicted ICU occupancy day 228 & -- & -- & 1.000 & -- \\
predicted ICU occupancy day 229 & -- & -- & 1.000 & -- \\
predicted ICU occupancy day 230 & -- & -- & 1.001 & -- \\
\end{longtable}
\endgroup

\paragraph*{S5 Table.}\label{S5_Table}
Detailed posterior agreement, predictive performance, and runtime metrics for all SBI configurations. For each inference setting, simulation budget, batch size, and epidemic window, the table reports parameter-wise agreement between SBI and MCMC posterior distributions measured using the first Wasserstein distance, Kullback--Leibler (KL) divergence, and symmetric KL divergence. Metrics are provided separately for all inferred epidemiological parameters and contact-rate change-point parameters. In addition, posterior predictive performance (RMSE), runtime statistics, number of failed runs, and simulation budgets are reported. Values are summarized as mean and standard deviation across 16 repeated SBI runs.

\end{document}